\documentclass[12pt]{article}
\usepackage{natbib}
\usepackage[hidelinks]{hyperref}
\usepackage{fullpage}
\usepackage{graphicx}
\usepackage{caption}
\usepackage{subcaption}
\usepackage{amsmath}
\usepackage{amssymb}
\usepackage{mathtools}
\usepackage{pgfplots}
\pgfplotsset{compat=1.17}

\title{A Sparse Coding Interpretation of Neural Networks and Theoretical Implications}

\author{Joshua Bowren\\
\large{Department of Computer Science, University of Miami,} \\
\large{Coral Gables, FL, USA}  \\
\large{jbowren@miami.edu}}
\date{}

\begin{document}

\maketitle

\let\thefootnote\relax\footnote{This work has been submitted to the IEEE for possible publication. Copyright may be transferred without notice, after which this version may no longer be accessible.}

\begin{abstract}
Neural networks, specifically deep convolutional neural networks, have achieved unprecedented performance in various computer vision tasks, but the rationale for the computations and structures of successful neural networks is not fully understood. Theories abound for the aptitude of convolutional neural networks for image classification, but less is understood about why such models would be capable of complex visual tasks such as inference and anomaly identification. Here, we propose a sparse coding interpretation of neural networks that have ReLU activation and of convolutional neural networks in particular. In sparse coding, when the model's basis functions are assumed to be orthogonal, the optimal coefficients are given by the soft-threshold function of the basis functions projected onto the input image. In a non-negative variant of sparse coding, the soft-threshold function becomes a ReLU. Here, we derive these solutions via sparse coding with orthogonal-assumed basis functions, then we derive the convolutional neural network forward transformation from a modified non-negative orthogonal sparse coding model with an exponential prior parameter for each sparse coding coefficient. Next, we derive a complete convolutional neural network without normalization and pooling by adding logistic regression to a hierarchical sparse coding model. Finally we motivate potentially more robust forward transformations by maintaining sparse priors in convolutional neural networks as well performing a stronger nonlinear transformation.
\end{abstract}

\section{Introduction}

Advances in deep neural networks, especially deep convolutional neural networks, have made neural networks one of the premier approaches to various computer vision tasks such as image classification \citep{krizhevsky:nips12,simonyan:arxiv14,szegedy:cvpr15,he:cvpr16}, image segmentation \citep{long:cvpr15}, and content generation \citep{goodfellow:nips14,radford:arxiv15}. Strangely, however, precisely why these networks have performed so well on these tasks is not fully understood. Recent work by \cite{cohen:nat20} explained how neural responses to image classses comprise manifolds that convolutional neural networks may render linearly-separable for a final classifer, but the specific improvements needed to mitigate issues like neural network fooling \citep{szegedy:arxiv13,nguyen:cvpr15} are not obvious. While the goal of a classic deep convolutional neural network is to maximize image classification accuracy, the mathematical intuition for why the specific model computations and topology of a convolutional neural network like Alexnet \citep{krizhevsky:nips12} would enable it to outperform all other models of its time on the ImageNet Large Scale Visual Recognition Challenge \citep{russakovsky:ijcv15} remains mysterious.

Over the years, many have attempted principled approaches to computer vision with varying success \citep{fukushima:ccnn82,fukushima:nn88,riesenhuber:natneuro99,lowe:ijcv04}, but the seemingly ad-hoc convolutional neural network attained unprecedented results in image classification \citep{krizhevsky:nips12}, and then in the other computer vision tasks previously mentioned. The machine learning and computer vision communities were left with a puzzling question: why would models (coupled with a classifier) based on good principles such as building image features \citep{lowe:ijcv04}, generating images from a function of latent variables \citep{bell:nc95,olshausen:nat96,hinton:sci06}, minimizing first-order dependencies between latent variables \citep{bell:nc95,hyvarinen:nc97}, making model neuron responses independent \citep{coen:plos12,coen:jov13}, estimating image distributions \citep{karklin:nc05,karklin:nat09}, and capturing properties of the visual system \citep{fukushima:ccnn82,riesenhuber:natneuro99} be vastly outperformed by an ad-hoc model. Perhaps the answer to this question is less obscure after understanding the history of the hierarchical vision models leading up to the convolutional neural network's prominence.

The early neocognitron perhaps provided much of the inspiration of modern convolutional neural networks. \cite{fukushima:nn88} incorporated network structures similar to convolutional layers and even had a rectification (ReLU) stage after integration of excitatory and inhibitory neurons. The neocognitron was more faithful to neural mechanisms like synaptic transmission than modern day neural networks, and the model was even able to classify digits to some extent. However, with the introduction of more advanced convolutional neural networks able to recognize digits \citep{lecun:ieee98}, interest in the neocognitron waned. Shortly after, interest in neural networks in general also decreased due to the lack of big datasets and the ability to train the deeper neural networks needed to process them. Some started exploring unsupervised neural networks in an effort to side-step the need for big datasets \citep{hinton:nc06,hinton:sci06,salakhutdinov:icml07}, and implementations of hierarchical unsupervised algorithms followed \citep{lee:nips07,le:icass13}. However, interest in neural networks did not resurge until the success of Alexnet \citep{krizhevsky:nips12} with the availability of larger datasets and faster hardware (GPUs). This innovation occured despite a seemingly supervised-only approach to computer vision.

Another model that seems to have influenced convolutional neural networks was the sparse coding model of \cite{olshausen:nat96} which increased in popularity with the discovery of basis functions resembling Gabor wavelets. After this discovery, the attention on sparse coding then turned to hierarchical sparse coding \citep{karklin:nc05,karklin:nat09}, and some of the deep neural networks of the coming years incorporated sparsity as an important principle of representation \citep{lee:nips07,le:icass13}. A promising direction of hierarchical sparse coding research was learning the model's basis functions with a loss function that takes into account all of the model parameters like in the work of \cite{zeiler:citeseer10}, but interest seemed to have waned at the time with the rise of deep learning, though there has been some resurgence of interest in non-convolutional formulations like that of \cite{boutin:plos21} that learn several parameters at the same time. Perhaps some of the inspiration of sparse coding in convolutional neural networks is apparent through the interpretation of sparse firing in convolutional neural networks via the selectivity of filters convolved over a natural image.

Convolutional sparse coding models followed the prominence of the convolutional neural network \citep{bristow:cvpr13,wohlberg:icassp14}. However, most convolutional sparse coding methods modified the model such that the original probabilistic interpretation was lost. These convolutional sparse coding methods learned sparse features maps for convolving with a set of filters, rather than finding a sparse linear combination of basis functions to reconstruct images. While the difference seems subtle, the change forced what were the model's basis functions to be filters for sparse feature maps, and the original probabilistic interpretation was lost. When stacked in layers, these convolutional sparse coding models are closer to a convolutional neural network with a regularized reconstruction loss function for each layer rather than a sparse coding model with input gathered from convolution. One advantage of this formulation was that redundant basis functions like simple translations of a Gabor wavelet were avoided because the model's dictionary elements were not assumed to be independent \citep{bristow:cvpr13}, an assumption made in sparse coding for convenience. However, the Laplacian prior over input images was lost in favor of a sparse feature map. Convolutional sparse coding seemed to learn more interesting features than traditional sparse coding, but the theoretical understanding of why such a model would be excellent for computer vision was not fully understood, and much less for convolutional neural networks.

After the initial deep learning research boom, various posited theories \citep{papyan:jmlr17,du:nips18,zhou:acha20,zhou:nn20} attempting to explain properties of convolutional neural networks and how they help solve computer vision problems. However, for many the main goal was to explain why these models would make for good image classifiers, rather than build a realistic generative model suitable for a classifier. One of these papers that differed from this focus was that of \cite{papyan:jmlr17}, where the authors noticed a connection between convolutional neural networks and a version sparse coding constrained to have non-negative responses called non-negative sparse coding \citep{hoyer:nnsp02}. In particular, \cite{papyan:jmlr17} brought to attention the fact that \emph{the nonlinearity of the non-negative soft-thresholding function (soft-thresholding function with all negative values set to zero) and the ReLU nonlinearity of a convolutional neural network are the same nonlinearity}. \cite{papyan:jmlr17} derived this solution via a non-negative form of the Sparse-Land model \citep{elad:tip06} with an L1 penalty and applying filters (not assumed to be orthogonal) to the image to derive the sparse coefficient vector, but the solution is equivalent to solving for the coefficients of non-negative sparse coding with one basis function, then repeating the solution for several coefficients. This connection of sparse coding to convolutional neural networks was striking and full of implications, but few have taken advantage of this important result; some of those who explored these implications have obtained noteworthy results \citep{sulam:tsp18,sulam:tpami19}.

In this work, we found the result of \cite{papyan:jmlr17} independently by solving the sparse inference problem of sparse coding (solving for the coefficients) from an orthogonal-only basis function version of the model. The result was found after accidentally assuming the basis functions to be orthogonal, an assumption referred to here as the \emph{orthogonality assumption}. This approach falls under a class of approaches known as \emph{orthogonal sparse coding}, described by \cite{schutze:ieeeci16} for the particular case of L0-regularized orthogonal sparse coding. The authors noted that solving for the coefficients of L0-regularized orthogonal sparse coding can be done with an exact solution, but the connection to convolutional neural networks was not clear because the solution was not a piecewise nonlinear function with a couple of conditions like that of the soft-thresholding function. Instead, the optimal coefficients were computed via a linear multiplication of the image by a matrix (the transpose of the basis function matrix) followed by setting several of the smallest magnitude entries of the product to zero based on the value of the regularization coefficient (a nonlinear transformation). Here, orthogonal sparse coding was performed with a L1 regularization penalty (the sparse inference problem being equivalent to LASSO, the usual formulation) instead of a L0 regularization penalty. The model is roughly equivalent to independent component analysis (ICA) with noise \citep{hyvarinen:ncomputing98,hyvarinen:nc99} except with an adjustable Laplacian hyperparameter (determining the sparse coding regularization coefficient along with the variance of the noise). The solution to this problem, perhaps unsurprisingly to some, turned out to be the soft-thresholding function. While this result on the surface seems trivial, it establishes the nonlinear nature of sparse coding, although intuition might imply that the linear generative model of sparse coding implies a linear solution. In fact, the true sparse coding transform contains a stronger implicit nonlinearity, the solution to non-orthogonal LASSO, because the orthogonal model reduces the generality of sparse coding.

In the light of the nonlinear nature of orthogonal sparse coding, and existing extensions to sparse coding like the non-negative variant of \cite{hoyer:nnsp02}, the connection to convolutional neural networks is apparent when comparing the soft-thresholding function and the ReLU function: the ReLU function is equivalent to the soft-thresholding function with all negative values set to zero, as \cite{papyan:jmlr17} noticed. Therefore, a model with the non-negative soft-thresholding function as its forward transformation, like the orthogonal version of non-negative sparse coding derived here, has the same nonlinearity as a convolutional neural network. We also show here that the input to the non-negative soft-thresholding function in non-negative orthogonal sparse coding is a matrix multiplied by the input image, so the forward transformation is equivalent to that of a convolutional neural network under some conditions (constant bias neuron weight over each set of weights equal to the regularization coefficient of non-negative orthogonal sparse coding). With this important derivation, we go on to show that \emph{the full forward transform of a convolutional neural network can be derived from a modified non-negative orthogonal sparse coding model}. To our knowledge, such a derivation has not been shown to date, and the implication is that the forward transformation of feed-forward multilayer neural networks with ReLU activation are stacks of a special form of non-negative orthogonal sparse coding. This contribution is an interesting new way to view certain neural networks, and convolutional neural networks in particular, but the connection to fully connected layers can also be made.

In this work, we also show that a complete convolutional neural network with fully connected layers (but not pooling or normalization layers) can be derived from a modified hierarchical sparse coding model. We show here that logistic regression can be incorporated into a hierarchical sparse coding model by a logistic regression layer on top of the last sparse coding layer and slightly changing the sparse coding derivation of \cite{olshausen:vr97}. Instead of adapting the model's basis functions to estimate the distribution of natural images, the basis functions (along with logistic regression weight matrices) can be adapted to estimate the conditional distribution of a target label given an input image with the label. This change gives the model's basis functions a cross-entropy loss function, just as a convolutional neural network. We then discuss adding principles of convolution to sparse coding by gathering the input patches via sliding a window over the input image. The process is equivalent to replacing the dot product of discrete convolution with sparse coding. The resulting model is a somewhat principled approach to image classification based on the advantages of sparse coding representations \citep{willshaw:nat69,kanerva:nasaames92} and the need to classify images with small error. However, there are several apparent issues under this sparse coding interpretation of convolutional neural networks.

First, the derivation of the convolutional neural network forward transform from the modified sparse coding model described here breaks down for positive bias weights. In order to maintain the prior distribution, the bias weights need to be constrained to be negative. Also, when the bias weights are close to zero, the prior distribution is hardly sparse and there is little effect of the prior. The magnitude of bias weights should be relatively large to benefit from the prior distribution. Second, the orthogonality assumption is made when deriving the sparse coding coefficients, but there is no reason to assume orthogonal basis functions in the probabilistic model. The general sparse inference problem contains a stronger nonlinearity than the orthogonal-only model, and we show here via an analogy to ridge regression the importance of the nonlinearity lost due to the orthogonality assumption. We also consider the implications of the L0-regularized sparse inference problem. Finally, we discuss how the feedforward arrangement of a convolutional neural network may not be sufficient to model the feedforward and feedback connections in the brain, and suggest a route for deriving an alternative transformation. We present all of these potential shortcomings of convolutional neural networks as tests for this theoretical framework in future work. If correct, the implications for neural networks could be significant. Better image models, perhaps ones that are resistant to fooling \citep{szegedy:arxiv13,nguyen:cvpr15}, have a prior image structure closer to the brain, not suffering from the artifactual effects of convolutional neural networks \citep{brendel:arxiv19}, may be derived by addressing the issues raised by this sparse coding interpretation.

\section{Sparse Inference}\label{sec:sparseinf}

\subsection{General Case}

The objective of the sparse inference problem of sparse coding is to find the optimal set of coefficients $\mathbf{\hat{a}}$ such that
\begin{align}
	\mathbf{\hat{a}} = \arg \min_\mathbf{a} \frac{1}{2} \left\Vert \mathbf{i} - \mathbf{\Phi} \mathbf{a} \right\Vert_2^2 + \lambda \left\Vert \mathbf{a} \right\Vert_1\label{eq:sparseinf}
\end{align}

\noindent under the generative model
\begin{align}
	\mathbf{i} = \mathbf{\Phi} \mathbf{a} + \boldsymbol{\nu}
\end{align}

\noindent where $\mathbf{i}$ is the input image, $\mathbf{\Phi}$ is the basis function matrix, $\mathbf{a}$ is the sparse coefficient vector, $\boldsymbol{\nu}$ is Gaussian noise, and $\lambda$ is the L1 regularization coefficient which comes from a Laplacian prior on the sparse coefficients $a_i$ \citep[see][]{olshausen:pmb02}. Since $\mathbf{\Phi}$ is constant for a set of coefficients $\mathbf{a}$, the loss function $f_{inf}$ can be written as a function of one variable:
\begin{align}
	f_{inf}(\mathbf{a}) &= \frac{1}{2} \left\Vert \mathbf{i} - \mathbf{\Phi} \mathbf{a} \right\Vert_2^2 + \lambda \left\Vert \mathbf{a} \right\Vert_1.
\end{align}

\noindent The subgradient $\nabla f_{inf}(\mathbf{a})$ with respect to $\mathbf{a}$ is
\begin{align}
	\nabla_\mathbf{a} f_{inf}(\mathbf{a}) &= \nabla_\mathbf{a} \left( \frac{1}{2} \left\Vert \mathbf{i} - \mathbf{\Phi} \mathbf{a} \right\Vert_2^2 + \lambda \left\Vert \mathbf{a} \right\Vert_1 \right) \\
	&= \mathbf{\Phi}^{\textrm{T}} \left( \mathbf{\Phi} \mathbf{a} - \mathbf{i} \right) + \nabla_{\mathbf{a}} \left( \lambda \left\Vert \mathbf{a} \right\Vert_1 \right) \\
	&= \mathbf{\Phi}^{\textrm{T}} \mathbf{\Phi} \mathbf{a} - \mathbf{\Phi}^{\textrm{T}} \mathbf{i} + \nabla_{\mathbf{a}} \left( \lambda \left\Vert \mathbf{a} \right\Vert_1 \right).
\end{align}

\noindent Note that the regular gradient of the L1 norm is undefined for any $a_i = 0$, but this issue can be avoided by optimizing each coefficient $a_i$ via its subdifferential set. The subdifferential set $\partial_{a_i} f_{inf}(\mathbf{a})$ for coefficient $a_i$ is given by
\begin{align}
	\partial_{a_i} f_{inf}(\mathbf{a}) = \left( \mathbf{\Phi}^{\textrm{T}} \mathbf{\Phi} \right)_{,i} \mathbf{a} - \mathbf{\Phi}_i^{\textrm{T}} \mathbf{i} + \lambda \partial_{a_i} \left| a_i \right|
\end{align}

\noindent where $\left(\mathbf{\Phi}^{\textrm{T}} \mathbf{\Phi}\right)_{,i}$ refers to the ith row of $\mathbf{\Phi}^{\textrm{T}} \mathbf{\Phi}$ and $\mathbf{\Phi}_i$ the ith column vector of $\mathbf{\Phi}$. The solution $\mathbf{\hat{a}}$ with components $\hat{a}_i$ is located at the point where the subdifferential set contains 0:
\begin{align}
	0 &\in \left( \mathbf{\Phi}^{\textrm{T}} \mathbf{\Phi} \right)_{,i} \mathbf{\hat{a}} - \mathbf{\Phi}_i^{\textrm{T}} \mathbf{i} + \lambda \partial_{a_i} \left| \hat{a}_i \right| \\
	\implies \left( \mathbf{\Phi}^{\textrm{T}} \mathbf{\Phi} \right)_{,i} \mathbf{\hat{a}} &\in \mathbf{\Phi}_i^{\textrm{T}} \mathbf{i} - \lambda \partial_{a_i} \left| \hat{a}_i \right|.\label{eq:solbad}
\end{align}

\noindent This problem does not have a general closed-form solution, but under certain conditions a closed-form solution exists. First consider the rows of the left-hand-side of equation \ref{eq:solbad}:
\begin{align}
	\begin{bmatrix}
		\left(\mathbf{\Phi}^{\textrm{T}} \mathbf{\Phi}\right)_{,1}^{\textrm{T}} \mathbf{\hat{a}} \\
		\left(\mathbf{\Phi}^{\textrm{T}} \mathbf{\Phi}\right)_{,2}^{\textrm{T}} \mathbf{\hat{a}} \\
		\vdots \\
		\left(\mathbf{\Phi}^{\textrm{T}} \mathbf{\Phi}\right)_{,n}^{\textrm{T}} \mathbf{\hat{a}}
	\end{bmatrix}
	&= 
	\begin{bmatrix}
		\mathbf{\Phi}_1^{\textrm{T}} \mathbf{\Phi}_1 \hat{a}_1 + \mathbf{\Phi}_1^{\textrm{T}} \mathbf{\Phi}_2 \hat{a}_2 + \dots \mathbf{\Phi}_1^{\textrm{T}} \mathbf{\Phi}_n \hat{a}_n \\
		\mathbf{\Phi}_2^{\textrm{T}} \mathbf{\Phi}_1 \hat{a}_1 + \mathbf{\Phi}_2^{\textrm{T}} \mathbf{\Phi}_2 \hat{a}_2 + \dots \mathbf{\Phi}_2^{\textrm{T}} \mathbf{\Phi}_n \hat{a}_n \\
		\vdots \\
		\mathbf{\Phi}_n^{\textrm{T}} \mathbf{\Phi}_1 \hat{a}_1 + \mathbf{\Phi}_n^{\textrm{T}} \mathbf{\Phi}_2 \hat{a}_2 + \dots \mathbf{\Phi}_n^{\textrm{T}} \mathbf{\Phi}_n \hat{a}_n
	\end{bmatrix} \\
	&=
	\begin{bmatrix}
		\hat{a}_1 + \mathbf{\Phi}_1^{\textrm{T}} \mathbf{\Phi}_2 \hat{a}_2 + \dots \mathbf{\Phi}_1^{\textrm{T}} \mathbf{\Phi}_n \hat{a}_n \\
		\mathbf{\Phi}_2^{\textrm{T}} \mathbf{\Phi}_1 \hat{a}_1 + \hat{a}_2 + \dots \mathbf{\Phi}_2^{\textrm{T}} \mathbf{\Phi}_n \hat{a}_n \\
		\vdots \\
		\mathbf{\Phi}_n^{\textrm{T}} \mathbf{\Phi}_1 \hat{a}_1 + \mathbf{\Phi}_n^{\textrm{T}} \mathbf{\Phi}_2 \hat{a}_2 + \dots \hat{a}_n
	\end{bmatrix}.
\end{align}

\noindent where the diagonal elements of $\mathbf{\Phi}^{\textrm{T}} \mathbf{\Phi}$ can be assumed to be all ones as the basis functions in sparse coding are usually constrained to be of unit-norm for optimization purposes \citep{olshausen:pmb02}.

\subsection{Orthogonal Case}

While the problem still does not have a general closed-form solution, if we make the assumption that the basis functions are uncorrelated (i.e. $\mathbf{\Phi}_i^{\textrm{T}} \mathbf{\Phi}_j = 0, \forall i \neq j$), then a closed-form solution exists. Note this is equivalent to orthogonal basis functions, so we refer to the model as \emph{orthogonal sparse coding}. This formulation is close to ICA with noise \citep[reconstruction loss, see][]{hyvarinen:ncomputing98,hyvarinen:nc99}, but with the addition of a sparsity control (hyperparameter determining the kurtosis of the Laplacian prior). However, $\mathbf{\Phi}$ technically need not be square because the only assumption is that the columns of $\mathbf{\Phi}$ (the basis functions) are orthogonal, not necessarily the rows. We continue by substituting $\mathbf{\Phi}_i^{\textrm{T}} \mathbf{\Phi}_j = 0, \forall i \neq j$ to find the solution $\mathbf{\hat{a}}^{orth}$:
\begin{align}
	\begin{bmatrix}
		\left(\mathbf{\Phi}^{\textrm{T}} \mathbf{\Phi}\right)_{,1}^{\textrm{T}} \mathbf{\hat{a}}^{orth} \\
		\left(\mathbf{\Phi}^{\textrm{T}} \mathbf{\Phi}\right)_{,2}^{\textrm{T}} \mathbf{\hat{a}}^{orth} \\
		\vdots \\
		\left(\mathbf{\Phi}^{\textrm{T}} \mathbf{\Phi}\right)_{,n}^{\textrm{T}} \mathbf{\hat{a}}^{orth}
	\end{bmatrix}
	= 
	\begin{bmatrix}
		\hat{a}^{orth}_1 \\
		\hat{a}^{orth}_2 \\
		\vdots \\
		\hat{a}^{orth}_n
	\end{bmatrix}
	\label{eq:orthassume}
	=
	\mathbf{\hat{a}}^{orth}.
\end{align}

\noindent Based on equations \ref{eq:solbad} and \ref{eq:orthassume} each problem now takes on the same form:
\begin{align}
	\hat{a}^{orth}_i &\in \mathbf{\Phi}_i^{\textrm{T}} \mathbf{i} - \lambda \partial \left| \hat{a}^{orth}_i \right|.
\end{align}

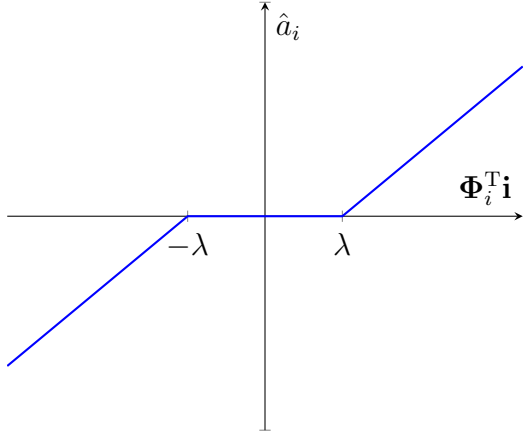
\begin{figure}
\centering
	\begin{tikzpicture}[
		declare function={
			func(\x)= (\x < -0.3) * (\x + 0.3)   +
					  and(\x >= -0.3, \x < 0.3) * (0)     +
					  (\x > 0.3) * (\x - 0.3)
			;
		}
		]
		\begin{axis}[
			axis x line=middle, axis y line=middle,
			ymin = -1, ymax = 1, ytick={-1,1}, yticklabels={}, ylabel=$\hat{a}_i$,
			xmin = -1, xmax = 1, xtick={-0.3,0.3}, xticklabels={$-\lambda$,$\lambda$}, xlabel=$\mathbf{\Phi}_i^{\textrm{T}} \mathbf{i}$,
			domain=-1:1,samples=101,
		]

		\addplot [blue,thick] {func(x)};
		\end{axis}
		\end{tikzpicture}
		\caption{\textbf{Soft-Threshold Function of $\mathbf{\Phi_i^{\textrm{T}}} \mathbf{i}$.} The function sets $a_i$ to 0 in the range $\left[-\lambda, \lambda\right]$, $\mathbf{\Phi}_i^{\textrm{T}} \mathbf{i} - \lambda$ in the range $\left[\lambda,\infty\right]$, and $\mathbf{\Phi}_i^{\textrm{T}} \mathbf{i} + \lambda$ in the range $\left[-\infty,-\lambda\right]$. In effect, the function drops the contributions of basis functions whose projection onto the input image $\mathbf{i}$ has a magnitude that falls below the threshold $\lambda$. Interestingly, the function is nonlinear, giving orthogonal sparse coding a nonlinear forward transformation.}
		\label{fig:soft-threshold}
\end{figure}

\noindent The derivative of the absolute value function is undefined at 0, so the problem can be solved by computing the subdifferential set when $a^{orth}_i > 0$, $a^{orth}_i = 0$, and $a^{orth}_i < 0$. The full derivation is given in \nameref{sec:appA}. The solution is given by the piecewise function
\begin{align}
	\hat{a}^{orth}_i = s\left(\mathbf{\Phi}_i^{\textrm{T}} \mathbf{i}\right) =
	\begin{cases} 
		\mathbf{\Phi}_i^{\textrm{T}} \mathbf{i} - \lambda & \mathbf{\Phi}_i^{\textrm{T}} \mathbf{i} > \lambda \\
		0 & \left| {\mathbf{\Phi}_i^{\textrm{T}} \mathbf{i}} \right| \leq \lambda \\
		\mathbf{\Phi}_i^{\textrm{T}} \mathbf{i} + \lambda & \mathbf{\Phi}_i^{\textrm{T}} \mathbf{i} < -\lambda,
	\end{cases}
	\label{eq:soft-threshold}
\end{align}

\noindent where $s\left(\mathbf{\Phi}_i^{\textrm{T}} \mathbf{i}\right)$ is the soft-threshold function of $\mathbf{\Phi}_i^{\textrm{T}} \mathbf{i}$. The solution is plotted in figure \ref{fig:soft-threshold}. The model may be thought of as projecting its basis functions onto the image, keeping only those projections whose magnitude is greater than the threshold $\lambda$, then adding or subtracting $\lambda$.

It is important to note the nonlinear nature of the solution in equation \ref{eq:soft-threshold}. Sparse coding is sometimes thought of as a linear model with a nonlinear forward transform, but substituting the exact solution for the special case into the objective for the derivation of the basis function matrix gives an idea for how non-linearity is incorporated into the generative model. The objective
\begin{align}\label{eq:orthscbasisloss}
	\mathbf{\hat{\Phi}} = &\arg \min_\mathbf{\Phi} \frac{1}{2} \sum_n \left\Vert \mathbf{i}_n - \mathbf{\Phi} \mathbf{s}\left(\mathbf{\Phi}^{\textrm{T}} \mathbf{i}_n\right) \right\Vert_2^2 + \lambda \left\Vert \mathbf{s}\left(\mathbf{\Phi}^{\textrm{T}} \mathbf{i}_n\right) \right\Vert_1 \\
	&\textrm{subject to } \mathbf{\Phi}^{\textrm{T}} \mathbf{\Phi} = \mathbf{I}
\end{align}

\noindent is a piecewise nonlinear function of $\mathbf{\Phi}$ where $\mathbf{s}(\mathbf{\Phi}^{\textrm{T}} \mathbf{i})$ is a vector with elements $s(\mathbf{\Phi}_i^{\textrm{T}} \mathbf{i})$, however sparse coding without the assumption that the basis functions are orthogonal may yield more nonlinear couplings in the objective function. Several methods can enforce the constraint $\mathbf{\Phi}^{\textrm{T}} \mathbf{\Phi} = \mathbf{I}$, but here we orthogonalize the basis functions after each gradient step with a SVD. This step takes the place of normalizing the basis functions in sparse coding. For a real $mxn$ $\mathbf{\Phi}$ the matrix can be decomposed via a SVD as 
\begin{align}\label{eq:svd}
	\mathbf{\Phi} &= \mathbf{U} \mathbf{\Sigma} \mathbf{V}^{\textrm{T}}
\end{align}

\noindent where $\mathbf{U}$ is an $mxm$ orthogonal matrix, $\mathbf{\Sigma}$ is an $mxn$ matrix with the singular values on the first n diagonals until the last rows of zeros, and $\mathbf{V}$ is an $nxn$ orthogonal matrix. Orthogonal basis functions can be achieved by setting the basis function matrix to
\begin{align}\label{eq:svdorth}
	\mathbf{\Phi}_{orth} &= \mathbf{\Phi} \mathbf{V} \mathbf{\Sigma}^{-1}.
\end{align}

\begin{figure}
	\centering
	\captionsetup[sub]{font=Large,labelfont={bf,sf}}
	\begin{subfigure}[normal]{0.40\columnwidth}
		\includegraphics[width=\columnwidth]{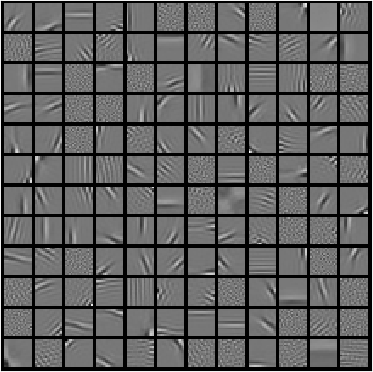}
		\caption{Orthogonal SC}
		\label{fig:orthscvsscvsicaorth}
	\end{subfigure}
	\begin{subfigure}[normal]{0.40\columnwidth}
		\includegraphics[width=\columnwidth]{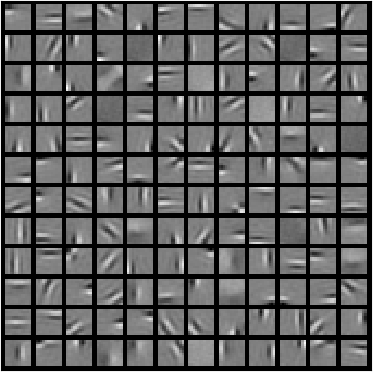}
		\caption{SC}
		\label{fig:orthscvsscvsicasc}
	\end{subfigure}
	\begin{subfigure}[normal]{0.40\columnwidth}
		\includegraphics[width=\columnwidth]{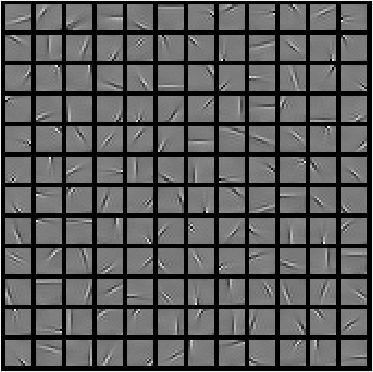}
		\caption{ICA}
		\label{fig:orthscvsscvsicaica}
	\end{subfigure}
	\caption{\textbf{Orthogonal Sparse Coding Solution and Other Models Trained on 12x12 Whitened Image Patches.} Both runs of sparse coding had a regularization coefficient of 0.1. (a) Orthogonal sparse coding basis functions contain Gabor-like filters and others that appear like Gabors with a 90-degree phase shift halfway through. Some basis functions appear as high spatial frequency noise. (b) Sparse coding basis functions contain Gabor-like basis functions that are generally smoother than the other two models. (c) ICA filters (basis functions) contain Gabor-like filters, but contain no high-spatial frequency basis functions like orthogonal sparse coding. }
	\label{fig:orthscvsscvsica}
\end{figure}

\begin{figure}
	\centering
	\captionsetup[sub]{font=Large,labelfont={bf,sf}}
	\begin{subfigure}[normal]{0.2\columnwidth}
		\includegraphics[width=\columnwidth]{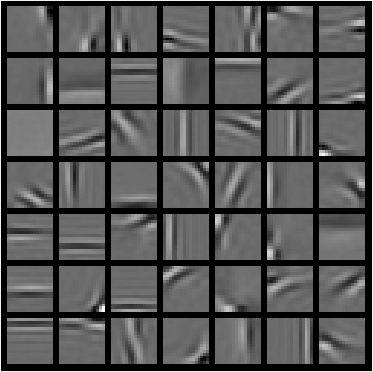}
		\caption{49}
	\end{subfigure}
	\begin{subfigure}[normal]{0.23\columnwidth}
		\includegraphics[width=\columnwidth]{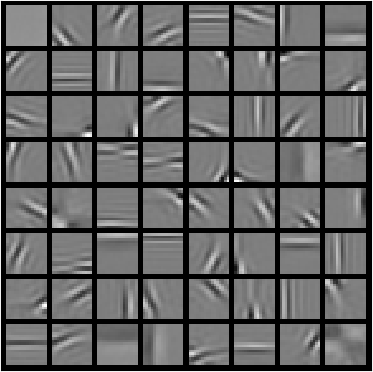}
		\caption{64}
	\end{subfigure}
	\begin{subfigure}[normal]{0.26\columnwidth}
		\includegraphics[width=\columnwidth]{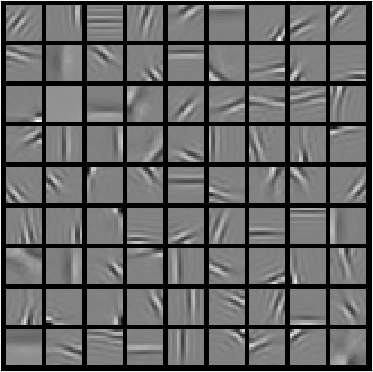}
		\caption{81}
	\end{subfigure}
	\begin{subfigure}[normal]{0.29\columnwidth}
		\includegraphics[width=\columnwidth]{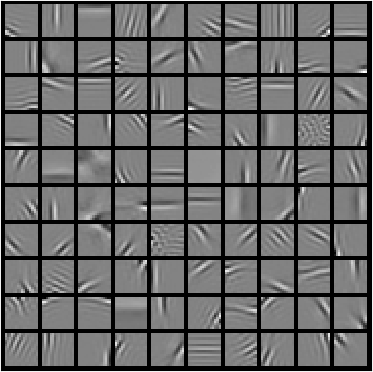}
		\caption{100}
	\end{subfigure}
	\begin{subfigure}[normal]{0.32\columnwidth}
		\includegraphics[width=\columnwidth]{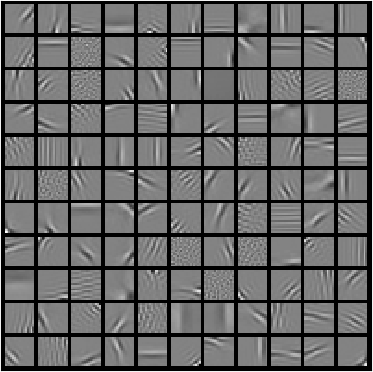}
		\caption{121}
		\label{fig:orthscbases121}
	\end{subfigure}
	\begin{subfigure}[normal]{0.35\columnwidth}
		\includegraphics[width=\columnwidth]{vis/vis144.png}
		\caption{144}
		\label{fig:orthscbases144}
	\end{subfigure}	
	\caption{\textbf{Orthogonal Sparse Coding With a Varying Number of Basis Functions.} The number below each image represents the number of basis functions. As the number of basis functions increases, the number of high spatial frequency noise basis functions also increases. High spatial frequency noise basis functions are absent for bases with a lower number of basis functions. Unique low spatial frequency basis functions seem to increase until the number of basis functions is greater than a little more than half the input dimension, after which high spatial frequency noise basis functions are introduced. By dropping high spatial frequency noise basis functions, this method may be viewed as a form of pooling.}
	\label{fig:orthscbases}
\end{figure}

The solution found for sparse coding (trained on whitened natural images with a regularization coefficient of 0.1; see figure \ref{fig:orthscvsscvsica}) with the constraint of orthogonal basis functions contains several high spatial frequency basis functions that probably do not contribute significantly to the model's representation. Also present are units reminiscent of the double-Gabor functions found by \cite{saremi:nc13} with a 90-degree phase-shift midway. The solution is shown next to that of regular sparse coding and ICA. The high spatial frequency basis functions are likely due to the orthogonality constraint; a vector space of $\mathbb{R}^n$ can have at most $n$ linearly-independent vectors, and the number of orthogonal basis functions needed to form the sparse reconstruction may be less. Indeed, as the number of basis functions learned by orthogonal sparse coding decreases, the number of apparent high spatial frequency basis functions decreases as shown in figure \ref{fig:orthscbases}. The number of non-high spatial frequency basis functions appears to be a little more than half of the input dimension. In this respect, orthogonal sparse coding may be thought of as a compressive \emph{pooling method} that discards non-orthogonal basis functions based on the Laplacian hyperparameter (or regularization coefficient). Such a method may be useful after the highly overcomplete transformation of visual area V1 in the brain \citep{olshausen:vn14}.

\section{Neural Network Interpretation}

\subsection{Non-Negative Sparse Coding}\label{sec:nnsc}

By examining the forward transform of orthogonal sparse coding in equation \ref{eq:soft-threshold}, we can see that the model may be interpreted as a single-layer neural network with the activation function
\begin{align}
	g_{\lambda}\left(x\right) &=
	\begin{cases} 
		x - \lambda & x > \lambda \\
		0 & \left|x\right| \leq \lambda \\
		x + \lambda & x < -\lambda.
	\end{cases}
	\label{eq:soft-threshold2}
\end{align}

\noindent The weight vectors of the neural network are the columns of $\mathbf{\Phi}$, there is no bias neuron, and the output neurons are $a_i$. The weights are adapted according to equation \ref{eq:orthscbasisloss}. Another formulation of sparse coding called non-negative sparse coding \citep{hoyer:nnsp02} constrains the sparse coefficients $a_i$ such that they are all non-negative. This is equivalent to assuming an exponential distribution over the sparse coefficients $a_i$ each with the same parameter $\mu$. The sparse coefficient vector $\mathbf{a}$ is obtained by minimizing the negative log-likelihood (equivalent to maximizing the likelihood) of $\mathbf{a}$ given the image $\mathbf{i}$. The likelihood function is proportional to the prior probability of the coefficients $P(\mathbf{a})$ and the probability of an image under the model $P(\mathbf{i} | \mathbf{a}, \mathbf{\Phi})$ given by
\begin{align}
	P(\mathbf{a}) &= \prod_i P(a_i) \\
	P(a_i)
	&= 
	\begin{cases} 
		\frac{1}{Z_{\mu}}\exp{\left( -\frac{a_i}{\mu} \right)} & a_i \geq 0 \\
		0 & \textrm{otherwise}
	\end{cases} \\
	P(\mathbf{i} | \mathbf{a}, \mathbf{\Phi}) &= \frac{1}{Z_{\sigma^2}}\exp{\left( -\frac{1}{2 \sigma^2} \Vert \mathbf{i} - \mathbf{\Phi} \mathbf{a} \Vert_2^2 \right)}.
\end{align}

\noindent \citep[see][]{olshausen:pmb02} where the coefficients $a_i$ are assumed to be independent, $\mu$ is the exponential prior parameter, $\sigma^2$ is the noise variance, and $Z_{\mu}$ and $Z_{\sigma^2}$ are the normalization constants for the two distributions. The likelihood function $g(\mathbf{a})$ is given by
\begin{align}
	P(\mathbf{a} | \mathbf{i}, \mathbf{\Phi}) \propto P(\mathbf{i} | \mathbf{a}, \mathbf{\Phi}) P(\mathbf{a}) &\propto g(\mathbf{a}) &= 
	\begin{cases} 
		\exp{\left( -\frac{1}{2 \sigma^2} \Vert \mathbf{i} - \mathbf{\Phi} \mathbf{a} \Vert_2^2 \right)} \prod_i \exp{\left( -\frac{a_i}{\mu} \right)} & a_i \geq 0, \forall i \\
		0 & \textrm{otherwise},
	\end{cases}
\end{align}

\noindent and the log-likelihood is
\begin{align}
	\log g(\mathbf{a}) =
	\begin{cases} 
		-\frac{1}{2 \sigma^2} \Vert \mathbf{i} - \mathbf{\Phi} \mathbf{a} \Vert_2^2 - \sum_i \frac{a_i}{\mu} & \forall a_i \geq 0 \\
		0 & \textrm{otherwise}.
	\end{cases}
\end{align}

\noindent Notice that the negative log-likelihood function is undefined for negative coefficients $a_i$, but is still needed to make the optimization problem easier. We can first solve the problem for the case $a_i \geq 0, \forall i$, then continue with the other case. The loss function for the problem for positive coefficients $a_i$ can be written as 
\begin{align}
	f_{nninf}(\mathbf{a}) = -\sigma^2 \log g(\mathbf{a}) = \frac{1}{2} \Vert \mathbf{i} - \mathbf{\Phi} \mathbf{a} \Vert_2^2 + \lambda \sum_i a_i, \forall a_i \geq 0
\end{align}

\noindent where $\lambda = \frac{\sigma^2}{\mu}$. Since the noise variance $\sigma^2$ is constant for a dataset, adjusting $\lambda$ adjusts the exponential prior parameter. The subgradient for $a_i > 0$ exists:
\begin{align}
	\nabla_a f_{nninf}(\mathbf{a})
	&= \nabla_a \left( \frac{1}{2} \Vert \mathbf{i} - \mathbf{\Phi} \mathbf{a} \Vert_2^2 + \lambda \sum_i a_i \right), \forall a_i > 0 \\
	&= \mathbf{\Phi}^{\textrm{T}} \left( \mathbf{\Phi} \mathbf{a} - \mathbf{i} \right) + \nabla_a \lambda \sum_i a_i \\
	&= \mathbf{\Phi}^{\textrm{T}} \mathbf{\Phi} \mathbf{a} - \mathbf{\Phi}^{\textrm{T}} \mathbf{i} + \nabla_a \lambda \sum_i a_i.
\end{align}

\noindent The gradient of the right most term is undefined when $\exists i : a_i \leq 0$, but the problem for all $a_i \geq 0$ can be solved with the subdifferential set (including $a_i = 0$). The subdifferential set $\partial_{a_i} f_{nninf}(\mathbf{a})$ for coefficient $a_i$ is given by
\begin{align}
	\partial f_{nninf}(\mathbf{a}) = \left( \mathbf{\Phi}^{\textrm{T}} \mathbf{\Phi} \right)_{,i} \mathbf{a} - \mathbf{\Phi}_i^{\textrm{T}} \mathbf{i} + \lambda \partial a_i, \forall a_i \geq 0.
\end{align}

\noindent Setting 0 as an element of the subdifferential set, the optimal $\mathbf{\hat{a}}$ can be solved with
\begin{align}
	0 &\in (\mathbf{\Phi}^{\textrm{T}} \mathbf{\Phi})_{,i} \mathbf{\hat{a}} - \mathbf{\Phi}_i^{\textrm{T}} \mathbf{i} + \lambda \partial \hat{a}_i, \forall \hat{a}_i \geq 0 \\
	\implies (\mathbf{\Phi}^{\textrm{T}} \mathbf{\Phi})_{,i} \mathbf{\hat{a}} &\in \mathbf{\Phi}_i^{\textrm{T}} \mathbf{i} - \lambda \partial \hat{a}_i, \forall \hat{a}_i \geq 0.
\end{align}

\noindent Applying the same orthogonality assumption in equation \ref{eq:orthassume}, we have
\begin{align}
	\hat{a}_i &\in \mathbf{\Phi}_i^{\textrm{T}} \mathbf{i} - \lambda \partial \hat{a}_i, \forall \hat{a}_i \geq 0.\label{eq:nnsol}
\end{align}

To solve the second case, i.e. when $\mathbf{a}$ has one or more negative coefficients, recall that we wish to maximize the likelihood function, but are optimizing the negative log-likelihood function for convenience. Note that the likelihood function has a constant value of 0 which is the minimum possible probability. Therefore, for any choice of negative $a_i$, we know that any other choice is at least as good. If we consider the value of the likelihood function when $\mathbf{a}$ is a vector of all zeros, we can clearly see that $f_{nninf}(0) = \exp{\left( -\frac{1}{2} \left\Vert \mathbf{i} \right\Vert_2^2 \right)}$ which is greater than zero, so the solution must have all non-negative $a_i$. Note that if $\mathbf{\Phi}_i^{\textrm{T}} \mathbf{i} < 0$, then both terms of the loss function decrease as $a_i$ decreases until the minimum of $a_i = 0$ is reached, so the solution for $\mathbf{\Phi}_i^{\textrm{T}} \mathbf{i} < 0$ is $a_i = 0$. Finally, equation \ref{eq:nnsol} can be solved by going through the cases $a_i > 0$, $a_i = 0$, and $a_i < 0$ which is derived in \nameref{sec:appB}. The solution is given by the equation
\begin{align}
	\hat{a}^{orth}_i = s_{nn,\lambda}(\mathbf{\Phi}_i^{\textrm{T}} \mathbf{i}) &=
	\begin{cases} 
		\mathbf{\Phi}_i^{\textrm{T}} \mathbf{i} - \lambda & \mathbf{\Phi}_i^{\textrm{T}} \mathbf{i} > \lambda \\
		0 & \textrm{otherwise}.
	\end{cases}\label{eq:nnsoli}
\end{align}

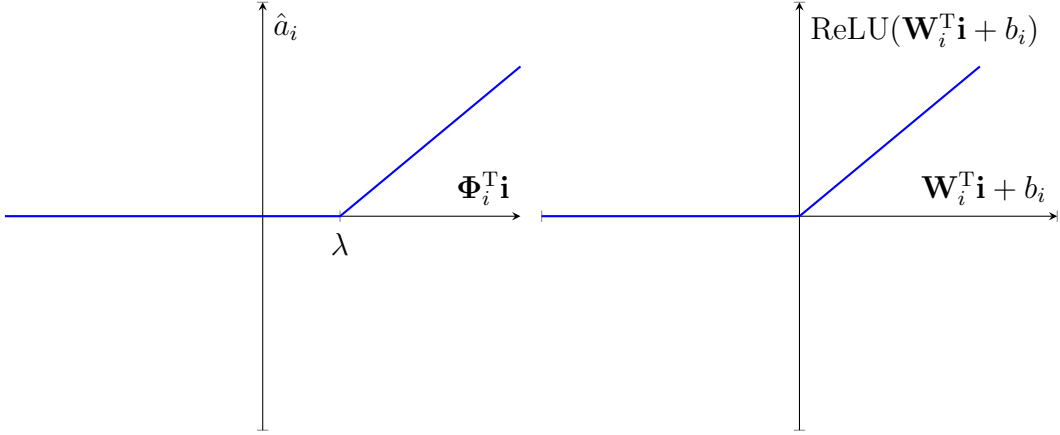
\begin{figure}
\centering
	\begin{tikzpicture}[
		declare function={
			func(\x)= (\x > 0.3) * (\x - 0.3)   +
					  (\x <= 0.3) * (0)
			;
		}
		]
		\begin{axis}[
			axis x line=middle, axis y line=middle,
			ymin = -1, ymax = 1, ytick={-1,1}, yticklabels={}, ylabel=$\hat{a}_i$,
			xmin = -1, xmax = 1, xtick={0.3}, xticklabels={$\lambda$}, xlabel=$\mathbf{\Phi}_i^{\textrm{T}} \mathbf{i}$,
			domain=-1:1,samples=101,
		]

		\addplot [blue,thick] {func(x)};
		\end{axis}
	\end{tikzpicture}
	\begin{tikzpicture}[
		declare function={
			func(\x)= (\x > 0) * (\x)   +
					  (\x <= 0) * (0)
			;
		}
		]
		\begin{axis}[
			axis x line=middle, axis y line=middle,
			ymin = -1, ymax = 1, ytick={-1,1}, yticklabels={}, ylabel=$\textrm{ReLU}(\mathbf{W}_i^{\textrm{T}} \mathbf{i} + b_i)$,
			xmin = -1, xmax = 1, xtick={-1,1}, xticklabels={}, xlabel=$\mathbf{W}_i^{\textrm{T}} \mathbf{i} + b_i$,
			domain=-1:1-0.3,samples=101,
		]

		\addplot [blue,thick] {func(x)};
		\end{axis}
	\end{tikzpicture}
		\caption{\textbf{Non-Negative Soft-Threshold Function of $\mathbf{\Phi}_i^{\textrm{T}} \mathbf{i}$ vs ReLU of $\mathbf{W}_i^{\textrm{T}} \mathbf{i} + b_i$.} The non-negative soft-threshold function is equivalent to a ReLU shifted to the right on the x-axis by $\lambda$. If the bias neuron weights are set to zero and $\lambda = 0$ (though it cannot be since $\sigma^2 > 0$), both orthogonal non-negative sparse coding and a convolutional layer of a convolutional neural network with a ReLU activation function compute the same output.}
		\label{fig:nnscvscnn}
\end{figure}

\noindent Going back to the neural network interpretation, we can view orthogonal non-negative sparse coding as a neural network with activation function
\begin{align}
	h_{\lambda}\left(x\right) &=
	\begin{cases} 
		x - \lambda & x > \lambda \\
		0 & x \leq \lambda
	\end{cases},
	\label{eq:soft-threshold3}
\end{align}

\noindent and weight vectors being the columns of $\mathbf{\Phi}$. Once again, there are no bias neurons, but the parameter $\lambda$ provides a shift for all the output neurons $a_i$. If we plot the orthogonal non-negative sparse coding solution from equation \ref{eq:nnsoli} as we did for orthogonal sparse coding, we can see that the solution resembles the forward transform of a convolutional layer of a convolutional neural network. The plot is shown next to that of a convolutional neural network in figure \ref{fig:nnscvscnn}. Notice how if $\lambda = 0$ (though it cannot be since $\sigma^2 > 0$) and the bias weights are all zero, then the models produce the same output, and as $\lambda$ approaches zero the models approach the same output. In other words, a convolutional neural network without bias weights is not viewed as sparse under this interpretation.

\subsection{Convolutional Neural Network Forward Transform Derivation}\label{sec:cnnftder}

If we modify the non-negative sparse coding formulation such that each sparse coefficient $a_i$ has its own exponential parameter $\mu_i$, then we get the modified likelihood function
\begin{align}
	P(\mathbf{a} | \mathbf{i}, \mathbf{\Phi}) \propto P(\mathbf{i} | \mathbf{a}, \mathbf{\Phi}) P(\mathbf{a}) &\propto g(\mathbf{a}) = 
	\begin{cases} 
		\exp{\left( -\frac{1}{2 \sigma^2} \Vert \mathbf{i} - \mathbf{\Phi} \mathbf{a} \Vert_2^2 \right)} \prod_i \exp{\left( -\frac{a_i}{\mu_i} \right)} & \forall a_i \geq 0 \\
		0 & \textrm{otherwise},
	\end{cases}
\end{align}

\noindent and the log-likelihood is given by
\begin{align}
	\log g(\mathbf{a}) = 
	\begin{cases} 
		-\frac{1}{2 \sigma^2} \Vert \mathbf{i} - \mathbf{\Phi} \mathbf{a} \Vert_2^2 - \sum_i \frac{a_i}{\mu_i} & a_i \geq 0, \forall i \\
		0 & \textrm{otherwise}.
	\end{cases}
\end{align}

\noindent Following the same procedure as before, we solve the problem first for $a_i \geq 0, \forall i$. The loss function for positive coefficients $a_i$ is given by
\begin{align}
	f_{nninf}(\mathbf{a}) = -\sigma^2 \log g(\mathbf{a}) = \frac{1}{2} \Vert \mathbf{i} - \mathbf{\Phi} \mathbf{a} \Vert_2^2 + \sum_i \lambda_i a_i, \forall a_i \geq 0
\end{align}

\noindent where $\lambda_i = \frac{\sigma^2}{\mu_i}$. We continue by minimizing the negative log-likelihood. The subgradient is
\begin{align}
	\nabla_a f_{nninf}(\mathbf{a}) &= \nabla_a \left( \frac{1}{2} \Vert \mathbf{i} - \mathbf{\Phi} \mathbf{a} \Vert_2^2 + \sum_i \lambda_i a_i \right), \forall a_i > 0 \\
	&= \mathbf{\Phi}^{\textrm{T}} \left( \mathbf{\Phi} \mathbf{a} - \mathbf{i} \right) + \nabla_{\mathbf{a}} \sum_i \lambda_i a_i \\
	&= \mathbf{\Phi}^{\textrm{T}} \mathbf{\Phi} \mathbf{a} - \mathbf{\Phi}^{\textrm{T}} \mathbf{i} + \nabla_{\mathbf{a}} \sum_i \lambda_i a_i
\end{align}

\noindent We continue to find the optimal coefficient vector $\mathbf{\hat{a}}$ by writing the subdifferential set
\begin{align}
	\partial_{a_i} f_{nninf}(\mathbf{a}) = \left( \mathbf{\Phi}^{\textrm{T}} \mathbf{\Phi} \right)_{,i} \mathbf{a} - \mathbf{\Phi}_i^{\textrm{T}} \mathbf{i} + \lambda_i \partial_{a_i} a_i, \forall a_i \geq 0,
\end{align}

\noindent and setting 0 as an element of the set:
\begin{align}
	0 &\in \left( \mathbf{\Phi}^{\textrm{T}} \mathbf{\Phi} \right)_{,i} \mathbf{\hat{a}} - \mathbf{\Phi}_i^{\textrm{T}} \mathbf{i} + \lambda_i \partial_{a_i} \hat{a}_i, \forall \hat{a}_i \geq 0 \\
	\implies \left( \mathbf{\Phi}^{\textrm{T}} \mathbf{\Phi} \right)_{,i} \mathbf{\hat{a}} &\in \mathbf{\Phi}_i^{\textrm{T}} \mathbf{i} - \lambda_i \partial_{a_i} \hat{a}_i, \forall \hat{a}_i \geq 0.
\end{align}

\noindent Applying the same orthogonality assumption we have
\begin{align}
	\hat{a}_i^{orth} &\in \mathbf{\Phi}_i^{\textrm{T}} \mathbf{i} - \lambda_i \partial_{a_i} \hat{a}_i, \forall a_i \geq 0.
\end{align}

\noindent With the same reasoning for non-negative sparse coding, we can see that the solution is never negative and that when $\mathbf{\Phi}_i^{\textrm{T}} \mathbf{i} - \lambda_i < 0$, the solution for $a_i$ is $\hat{a}_i = 0$. The solution takes the form
\begin{align}
	\hat{a}^{orth}_i = s_{nn,\lambda_i}(\mathbf{\Phi}_i^{\textrm{T}} \mathbf{i}) &=
	\begin{cases} 
		\mathbf{\Phi}_i^{\textrm{T}} \mathbf{i} - \lambda_i & \mathbf{\Phi}_i^{\textrm{T}} \mathbf{i} > \lambda_i \\
		0 & \textrm{otherwise}.
	\end{cases}\label{eq:nnsolcnn}
\end{align}

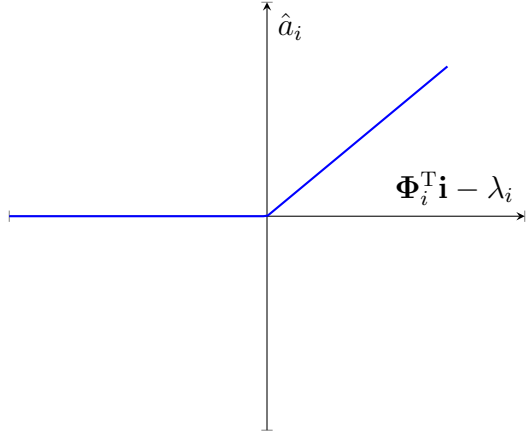
\begin{figure}
\centering
	\begin{tikzpicture}[
		declare function={
			func(\x)= (\x > 0) * (\x - 0)   +
					  (\x <= 0) * (0)
			;
		}
		]
		\begin{axis}[
			axis x line=middle, axis y line=middle,
			ymin = -1, ymax = 1, ytick={-1,1}, yticklabels={}, ylabel=$\hat{a}_i$,
			xmin = -1, xmax = 1, xtick={-1,1}, xticklabels={}, xlabel=$\mathbf{\Phi}_i^{\textrm{T}} \mathbf{i} - \lambda_i$,
			domain=-1:1-0.3,samples=101,
		]

		\addplot [blue,thick] {func(x)};
		\end{axis}
	\end{tikzpicture}
	\caption{\textbf{Non-Negative Soft-Threshold Function Written As Function of $\mathbf{\Phi}_i^{\textrm{T}} \mathbf{i} - \lambda_i$ With Exponential Parameters $\lambda_i$.} The non-negative soft-threshold function written as a function of $\mathbf{\Phi}_i^{\textrm{T}} \mathbf{i} - \lambda_i$ is equivalent to a ReLU.}
	\label{fig:cnn}
\end{figure}

While the solution can be written as equation \ref{eq:nnsolcnn}, the activation function may also be written as a function of $\mathbf{\Phi}_i^{\textrm{T}} \mathbf{i} - \lambda_i$; written this way the solution is the ReLU function of $\mathbf{\Phi}_i^{\textrm{T}} \mathbf{i} - \lambda_i$. The solution is plotted in figure \ref{fig:cnn}. \emph{The forward transform of this model is exactly the same as a convolutional layer of a convolutional neural network.} However, the probabilistic interpretation breaks down when any $\lambda_i$ is chosen to be negative because the Gaussian noise variance $\sigma^2$ and exponential prior parameter $\mu_i$ must both be positive. Others have also found the convolutional neural network forward transform by related sparse coding methods \citep{fawzi:ijcv15,papyan:jmlr17}. \cite{papyan:jmlr17}, for example, solved the L1 version of the Sparse-Land model \citep{elad:tip06} with linear filters applied to the image to derive the sparse coefficients instead of assuming orthogonal basis functions in sparse coding. While this method is closely related, the result here is unique. The loss function of this model adapts its weights to reconstruct input images rather than maximize image classification accuracy like a complete convolutional neural network:
\begin{align}\label{eq:orthscvariedlambda}
	\mathbf{\hat{\Phi}} = &\arg \min_{\mathbf{\Phi}} \frac{1}{2} \sum_n \left\Vert \mathbf{i}_n - \mathbf{\Phi} \mathbf{ReLU}\left(\mathbf{\Phi}^{\textrm{T}} \mathbf{i}_n - \boldsymbol{\lambda}\right) \right\Vert_2^2 + \boldsymbol{\lambda}^{\textrm{T}} \mathbf{ReLU}\left(\mathbf{\Phi}^{\textrm{T}} \mathbf{i}_n - \boldsymbol{\lambda}\right) \\
	&\textrm{subject to } \mathbf{\Phi}^{\textrm{T}} \mathbf{\Phi} = \mathbf{I}.
\end{align}

\noindent The weights of this model are the columns of $\mathbf{\Phi}$ and bias weights $-\boldsymbol{\lambda}$. The model also has a ReLU activation function.

It is important to note that all of the sparse coding models mentioned in this paper can be stacked in layers with one global loss function. This last model when stacked in layers is equivalent to a convolutional neural network without pooling or normalization, a different loss function (image reconstruction), and no classifiers trained on top. In this interpretation, the difference between the forward transformation of non-negative sparse coding and a convolutional neural network is that a convolutional neural network assigns a different exponential parameter $\mu_i$ to each sparse coefficient $a_i$ and non-negative sparse coding has one global exponential parameter $\mu$ for all sparse coefficients $a_i$. Also, a convolutional neural network chooses its parameters $\lambda_i$ such that they minimize its cross-entropy loss function along with its weights $\mathbf{\Phi}^{\textrm{T}}$ whereas the hyperparameter $\lambda$ in non-negative sparse coding is a positive value typically chosen before training. Interestingly, this last model and a convolutional neural network are similar to the second layer of the hierarchical sparse coding model of \cite{karklin:nc05} in that they all assign a different value $\lambda_i$ for every sparse coefficient $a_i$. However, for this last model the values of $\lambda_i$ are chosen before training, for a convolutional neural network they are explicitly learned via minimizing its loss function, and in the model of \cite{karklin:nc05} they are implicitly learned by training the model. The second layer of the model of \cite{karklin:nc05} interprets the values $\lambda_i$ as the variance of each input and generates them a nonlinear function of its basis functions and sparse coefficients.

\section{Stacking Generative and Discriminative Models}

\subsection{Sparse Coding}\label{sec:stacksparse}

\begin{figure}
	\centering
	\captionsetup[sub]{font=large,labelfont={bf,sf}}
	\begin{subfigure}[normal]{0.49\columnwidth}
		\centering
		\includegraphics[height=0.6\columnwidth]{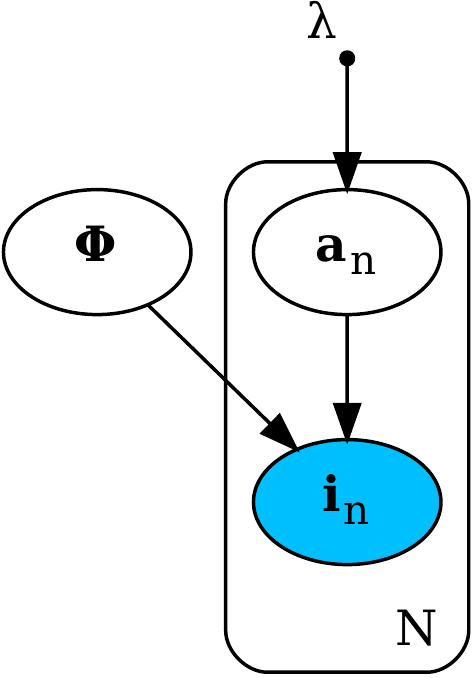}
		\caption{}
		\label{fig:genmodelsa}
	\end{subfigure}
	\begin{subfigure}[normal]{0.49\columnwidth}
		\centering
		\includegraphics[height=0.6\columnwidth]{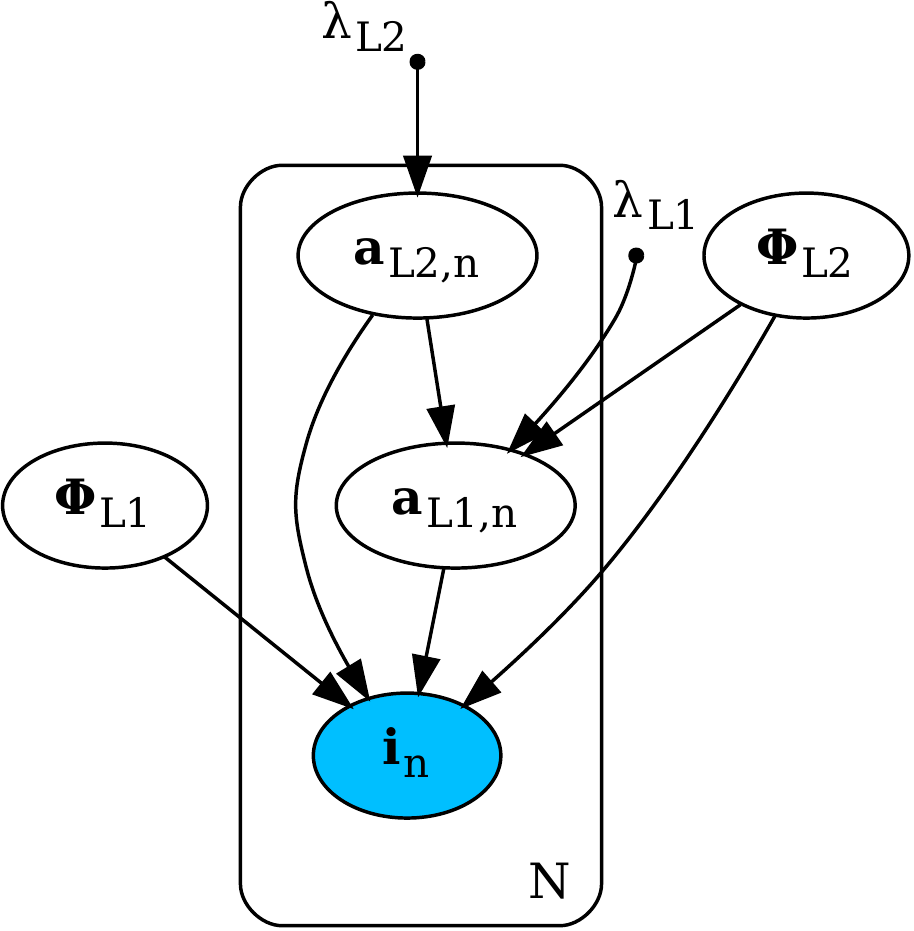}
		\caption{}
		\label{fig:genmodelsb}
	\end{subfigure}
	\caption{\textbf{Generative Sparse Coding Models.} Graphs for the joint distributions of two sparse coding models. $N$ is the number of training examples. The subscript $n$ denotes the training example. If applicable, the subscript $n$ is preceded by the layer (L1 or L2) followed by a comma. (a) Sparse coding with a factorization of its joint distribution such that $P(\mathbf{a} | \mathbf{i}, \mathbf{\Phi}, \mathbf{\lambda}) \propto P(\mathbf{i} | \mathbf{a}, \mathbf{\Phi})P(\mathbf{a} | \mathbf{\lambda})$. (b) Two-layer sparse coding capturing more dependence structure with a factorization of its joint distribution such that $P(\mathbf{a}_{\textrm{L1}}, \mathbf{a}_{\textrm{L2}} | \mathbf{i}, \mathbf{\Phi}_{\textrm{L1}}, \mathbf{\Phi}_{\textrm{L2}}, \mathbf{\lambda}_{\textrm{L1}}, \mathbf{\lambda}_{\textrm{L2}}) \propto P(\mathbf{i} | \mathbf{a}_{\textrm{L1}}, \mathbf{a}_{\textrm{L2}}, \mathbf{\Phi}_{\textrm{L1}}, \mathbf{\Phi}_{\textrm{L2}})P(\mathbf{a}_{\textrm{L1}}, \mathbf{a}_{\textrm{L2}} | \mathbf{\Phi}_{\textrm{L1}}, \mathbf{\Phi}_{\textrm{L2}}, \lambda_{\textrm{L1}})$. This model requires specifying the prior distribution $P(\mathbf{a}_{\textrm{L1}}, \mathbf{a}_{\textrm{L2}})$.}
	\label{fig:genmodels}
\end{figure}

Sparse coding models can be stacked in layers or combined with other generative models to build hierarchical sparse coding models \citep{karklin:nc05,karklin:nat09,lee:nips07}. The simplest realization is a two-layer sparse coding model. Here, the coefficients are modeled by exponential distributions instead of Laplacians to keep the neural network interpretation discussed previously. The first layer of the model proceeds as usual: the coefficients $\mathbf{a}_{\textrm{L1}}$ are chosen by maximizing $P(\mathbf{a}_{\textrm{L1}} | \mathbf{i}, \mathbf{\Phi}_{\textrm{L1}})$ according to:
\begin{align}\label{eq:sceqs}
	\mathbf{i} &= \mathbf{\Phi}_{\textrm{L1}} \mathbf{a}_{\textrm{L1}} \\
	P(\mathbf{a}_{\textrm{L1}}) &= \prod_i P(a_{\textrm{L1},i}) \\
	P(a_{\textrm{L1},i})
	&= 
	\begin{cases} 
		\frac{1}{Z_{\mu_{\textrm{L1}}}}\exp{\left( -\frac{a_{\textrm{L1},i}}{\mu_{\textrm{L1}}} \right)} & a_{\textrm{L1},i} \geq 0 \\
		0 & \textrm{otherwise}
	\end{cases} \\
	P(\mathbf{i} | \mathbf{a}_{\textrm{L1}}, \mathbf{\Phi}_{\textrm{L1}}) &\propto \left( \frac{1}{2} \Vert \mathbf{i} - \mathbf{\Phi}_{\textrm{L1}} \mathbf{a}_{\textrm{L1}} \Vert_2^2 + \lambda_{\textrm{L1}} \sum_i a_{\textrm{L1},i} \right),
\end{align}

\noindent where $\mathbf{\Phi}_{\textrm{L1}}$ and $\mathbf{a}_{\textrm{L1}}$ are model parameters and $\mu_{\textrm{L1}}$ and $\sigma^2_{\textrm{L1}}$ (noise variance written via $\lambda_{\textrm{L1}}$ implicitly as in section \ref{sec:nnsc}) are hyperparameters chosen to enforce sparseness in $\mathbf{a}_{\textrm{L1}}$ for the first layer, layer L$1$. In practice, only the regularization coefficient $\lambda_{Ln} = \frac{\sigma^2_{Ln}}{\mu_{Ln}}$ is needed to enforce sparsity, so it is the only parameter maintained in figure \ref{fig:genmodels}. In order to derive a convolutional neural network in the next section, the next layer, layer L$2$, considers its input vector $\mathbf{a}_{\textrm{L1}}$ as a constant variable like the input image $\mathbf{i}$ in layer L$1$, so it derives its coefficients $\mathbf{a}_{\textrm{L2}}$ by maximizing $P(\mathbf{a}_{\textrm{L2}} | \mathbf{a}_{\textrm{L1}}, \mathbf{\Phi}_{\textrm{L2}})$ according to:
\begin{align}\label{eq:sc2eqs}
	\mathbf{a}_{\textrm{L1}} &= \mathbf{\Phi}_{\textrm{L2}} \mathbf{a}_{\textrm{L2}} \\
	P(\mathbf{a}_{\textrm{L2}}) &= \prod_i P(a_{\textrm{L2},i}) \\
	P(a_{\textrm{L2},i})
	&= 
	\begin{cases} 
		\frac{1}{Z_{\mu_{\textrm{L2}}}}\exp{\left( -\frac{a_{\textrm{L2},i}}{\mu_{\textrm{L2}}} \right)} & a_{\textrm{L2},i} \geq 0 \\
		0 & \textrm{otherwise}
	\end{cases} \\
	P(\mathbf{a}_{\textrm{L1}} | \mathbf{a}_{\textrm{L2}}, \mathbf{\Phi}_{\textrm{L2}}) &\propto \left( \frac{1}{2} \Vert \mathbf{a}_{\textrm{L1}} - \mathbf{\Phi}_{\textrm{L2}} \mathbf{a}_{\textrm{L2}} \Vert_2^2 + \lambda_{\textrm{L2}} \sum_i a_{\textrm{L2},i} \right),
\end{align}

\noindent where $\mathbf{\Phi}_{\textrm{L2}}$ and $\mathbf{a}_{\textrm{L2}}$ are model parameters of layer L$2$ and $\mu_{\textrm{L2}}$ and $\sigma^2_{\textrm{L2}}$ (again, the noise variance written implicitly via $\lambda_{\textrm{L2}}$) are the hyperparameters of layer L$2$. This model is equivalent to performing single-layer sparse coding twice with the independence assumptions illustrated by the dependency graph in figure \ref{fig:genmodelsa}. If we continue as one-layer of sparse coding \citep{olshausen:vr97}, we wish to approximate the true distribution of natural images $P^*(\mathbf{i})$ and the supposed unknown true distribution $P^*(\mathbf{a}_{\textrm{L1}}$) (though we know the prior) with the distribution of images under the model $P(\mathbf{i} | \mathbf{\Phi}_{\textrm{L1}})$ and the distribution of $\mathbf{a}_{\textrm{L1}}$ under the model $P(\mathbf{a}_{\textrm{L1}} | \mathbf{\Phi}_{\textrm{L2}})$ respectively. These distributions are approximated by choosing the basis function matrices to minimize the KL-divergence of the true distributions with the distributions under the model. Note, this is equivalent to maximizing the log-likelihood of the inputs given the the basis function matrix of the layer being computed \citep{olshausen:vr97}. The assumptions in this model make it easy to extend to any arbitrary number of layers simply by adding more applications of sparse coding.

From a sparse coding standpoint, some of the issues with this approach include: potentially important dependencies between model parameters are not captured, each layer is treated as its own model so knowledge of certain model parameters are unused, and each set of sparse coefficients only reconstructs inputs from the previous layer. The last issue was addressed by \citep{zeiler:citeseer10} in a convolutional sparse coding model, and reconstruction loss terms that force each set of coefficients to reconstruct the input image can trivially be added to sparse coding. Recall that the product of two Gaussian probability distributions functions is also Gaussian. The reconstruction loss terms can be achieved by changing the generative model to
\begin{align}
	\mathbf{i} &= a \mathbf{\Phi}_{\textrm{L1}} \mathbf{a}_{\textrm{L1}} + b \mathbf{\Phi}_{\textrm{L1}} \mathbf{\Phi}_{\textrm{L2}} \mathbf{a}_{\textrm{L2}} + \boldsymbol{\nu}.
\end{align}

\noindent where the coefficients $a$ and $b$ choose the importance of the two reconstruction loss terms and implicitly define the noise variance of the two Gaussians along with the variance of $\boldsymbol{\nu}$. A more advanced hierarchical sparse coding model incorporating this dependency and dependencies between the sparse coefficients is illustrated in figure \ref{fig:genmodelsb}. Another work that addressed the issues present in performing sparse coding several times was that of \cite{boutin:plos21} where sets of coefficients were optimized taking into account the reconstructions of other layers. Hierarchical sparse coding models remain a promising open area of research.

\subsection{Adding Logistic Regression}\label{sec:addlogreg}

Another method of choosing the basis function matrix $\mathbf{\Phi}$ in a sparse coding model is to add a logistic regression model on top of the sparse coding model with a dependency between the model parameters $\mathbf{\Phi}$ and $\lambda$ (the second being a hyperparameter), and a one-of-K target vector $\mathbf{t}$ corresponding to image $\mathbf{i}_{\mathbf{t}}$ from which a subset of input image patches $\mathbf{i}$ are drawn. A dependency is also added between the target vector $\mathbf{t}$ and the input image $\mathbf{i}_{\mathbf{t}}$ with known label $\mathbf{t}$. This description of sparse coding with the addition of logistic regression only is limited in that its logistic regression classes are only associated with the input image patch $\mathbf{i}$ and not the input image $\mathbf{i}_{\mathbf{t}}$ from which the patch was drawn, but this can be alleviated by adding convolution principles to gather larger patches for subsequent layers as discussed in section \ref{sec:sparseconv}. For this section, the input image patch $\mathbf{i}$ may be taken to be the complete image $\mathbf{i}_{\mathbf{t}}$ with the corresponding label $\mathbf{t}$.

Logistic regression computes the probability that an input belongs to several classes as some function $\sigma$ (such as softmax or a sigmoid) of a weighted sum of the input values \citep{bishop:prml06}. The model can be extended to multiple target values \citep[known as multinomial logistic regression;][]{bishop:prml06} via multiple outputs as in a neural network by incorporating a matrix of weights $\mathbf{W}$ and a bias neuron weight vector $\mathbf{b}$. The output probability vector $\mathbf{y}$ is given by
\begin{align}\label{eq:logreg}
	\mathbf{y} = \sigma(\mathbf{W}\mathbf{x} + \mathbf{b}),
\end{align}

\noindent where $y_i$ indicates the probability the input vector $\mathbf{x}$ belongs to class $i$. With the matrix of target vectors $\mathbf{T}$, the loss function for the output $y_i$ can be shown to be the cross entropy loss function of neural networks \citep{bishop:prml06}:
\begin{align}\label{eq:logregloss}
	f_{logreg}(\mathbf{T}, \mathbf{W}, \mathbf{b}) = -\log P(\mathbf{T} | \mathbf{W}, \mathbf{b}) = -\sum_i \sum_j t_{i,j} \ln y_{i, j}.
\end{align}

Considering the two-layer sparse coding model with hyperparameters $\lambda_{\textrm{L1}}$ and $\lambda_{\textrm{L2}}$, we can add a dependency to the basis function matrix, the regularization coefficients, and the input image on the target values, and the sparse coding responses $a_i$ can serve as inputs to the logistic regression model. Importantly, the model does not derive its basis functions by minimizing the KL-divergence of the true distribution of images $P^*(\mathbf{i})$ and distribution of images under the model $P(\mathbf{i} | \mathbf{\Phi}_{\textrm{L1}}, \mathbf{\Phi}_{\textrm{L2}}, \lambda_{\textrm{L1}}, \lambda_{\textrm{L2}})$. Instead, the basis functions are chosen to minimize the KL-divergence of the true conditional distribution of a target vector given one of its input images $P^*(\mathbf{t} | \mathbf{i}_{\mathbf{t}})$ and the distribution under the model $P(\mathbf{t} | \mathbf{W}, \mathbf{b}, \mathbf{\Phi}_{\textrm{L1}}, \mathbf{\Phi}_{\textrm{L2}}, \lambda_{\textrm{L1}}, \lambda_{\textrm{L2}}, \mathbf{i}_{\mathbf{t}})$. The KL-divergence is given by 
\begin{align}\label{eq:kldiv}
	KL(P^* \lVert P) &=  \int P^*(\mathbf{t} | \mathbf{i}_{\mathbf{t}})\log\frac{P^*(\mathbf{t} | \mathbf{i}_{\mathbf{t}})}{P(\mathbf{t} | \theta, \mathbf{i}_{\mathbf{t}})}d\mathbf{t} \\
	&= \int P^*(\mathbf{t} | \mathbf{i}_{\mathbf{t}})\log P^*(\mathbf{t} | \mathbf{i}_{\mathbf{t}}) d\mathbf{t} - \int P^*(\mathbf{t} | \mathbf{i}_{\mathbf{t}})\log P(\mathbf{t} | \theta, \mathbf{i}_{\mathbf{t}}) d\mathbf{t} \\
	&= C + \langle -\log P(\mathbf{t} | \theta, \mathbf{i}_{\mathbf{t}}) \rangle_{P^*(\mathbf{t} | \mathbf{i}_{\mathbf{t}})}.
\end{align}

\noindent where $C$ is a constant with respect to the model parameters in the set $\{\mathbf{W}$, $\mathbf{b}$, $\mathbf{\Phi}_{\textrm{L1}}$, $\mathbf{\Phi}_{\textrm{L2}}$, $\lambda_{\textrm{L1}}$, $\lambda_{\textrm{L2}}\}$ denoted by $\theta$. Minimizing the KL-divergence of the true conditional distribution and that under the model is the same as minimizing the entropy in the target labels under the model:
\begin{align}\label{eq:kldivmin}
	\arg \min_{\theta} KL(P^* \lVert P) &= \arg \min_{\theta} C + \langle -\log P(\mathbf{t} | \theta, \mathbf{i}_{\mathbf{t}}) \rangle_{P^*(\mathbf{t} | \mathbf{i}_{\mathbf{t}})} \\
	&= \arg \min_{\theta} \langle -\log P(\mathbf{t} | \theta, \mathbf{i}_{\mathbf{t}}) \rangle_{P^*(\mathbf{t} | \mathbf{i}_{\mathbf{t}})}
\end{align}

\noindent For a finite set of observed target vectors $\mathbf{t}_n$ with corresponding input images $\mathbf{i}_{\mathbf{t}_n}$ we can approximate the entropy in $\mathbf{t}$ under the model with a sum and write the loss function for sparse coding with logistic regression as 
\begin{align}\label{eq:sclogregloss}
	f_{sclogreg}(\mathbf{T}, \theta, \{\mathbf{i}_{\mathbf{t}_n}\}) &= -\sum_n \log{P(\mathbf{t}_n | \theta, \mathbf{i}_{\mathbf{t}_n})} \\
	&= -\sum_n \log{\prod_m y_{n,m}^{t_{n,m}}} \\
	&= -\sum_n \sum_m t_{n,m} \log{y_{n,m}},
\end{align}

\noindent where $y_{n,m}$ is the mth component of the nth probability vector $\mathbf{y}_n$ computed with the second layer sparse coding coefficients as inputs. Notice that if we change the hyperparameters $\lambda_{\textrm{L1}}$ and $\lambda_{\textrm{L2}}$ to learnable parameter vectors $\boldsymbol{\lambda}_{\textrm{L1}}$ and $\boldsymbol{\lambda}_{\textrm{L2}}$ the forward transform for the sparse coding layers becomes a ReLU of the weight vectors times the input plus the learnable bias weights $-\boldsymbol{\lambda}_{\textrm{L}K}$ for layer $K$ (see section \ref{sec:cnnftder}). The multinomial logistic regression layer forward transformation is given by equation \ref{eq:logreg} with $\mathbf{x} = \mathbf{a}_{\textrm{L2}}$. More multinomial logistic regression layers can be added to improve the final classifier. \emph{Interestingly, this model is equivalent to a neural network without convolution, normalization, or pooling.} We will see next that the main idea behind convolution (sliding window over input) can easily be integrated into the model.

\subsection{Convolution}\label{sec:sparseconv}

Early attempts to perform hierarchical sparse coding learned one set of sparse coefficients for each set in the preceding layer \citep{karklin:nc05,lee:nips07}. However, this method makes consolidating image features across input image patches close in space problematic because each second layer neuron coded for only one input patch. Convolutional neural networks \citep{lecun:hbtnn95} and their recent prominence \citep{krizhevsky:nips12} bring to attention a method of consolidating image features across space (patches). In a convolutional neural network, higher level convolutional filters extend across the responses of multiple lower level convolutional filters and consolidates their responses with a dot product. This idea can also be incorporated in sparse coding for consolidation of image features across space.

\begin{figure}
	\centering
	\captionsetup[sub]{font=normalsize}
	\begin{subfigure}[normal]{0.45\columnwidth}
		\includegraphics[width=\columnwidth]{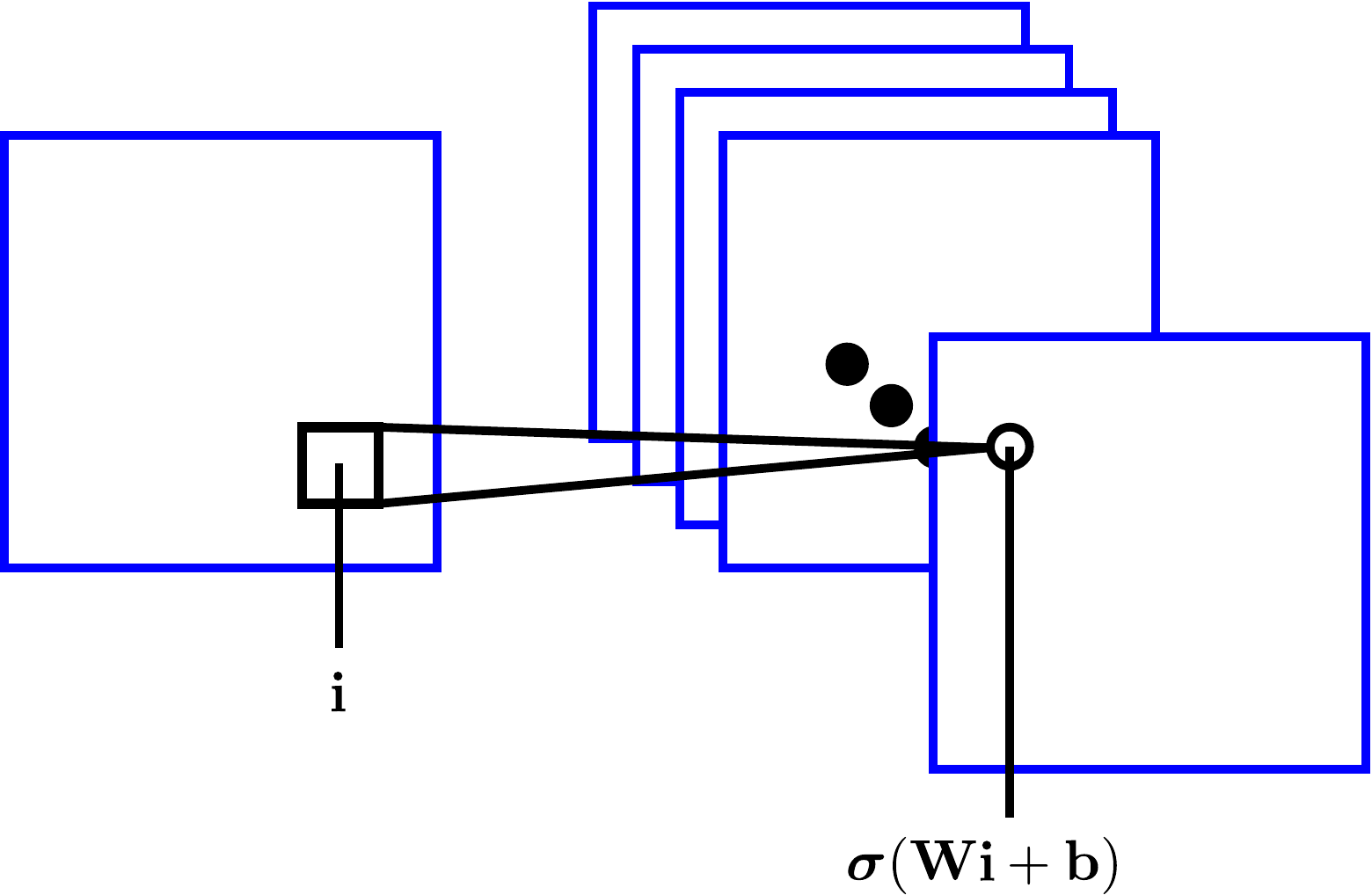}
		\caption{Convolution with ReLU}
	\end{subfigure}
	\begin{subfigure}[normal]{0.45\columnwidth}
		\includegraphics[width=\columnwidth]{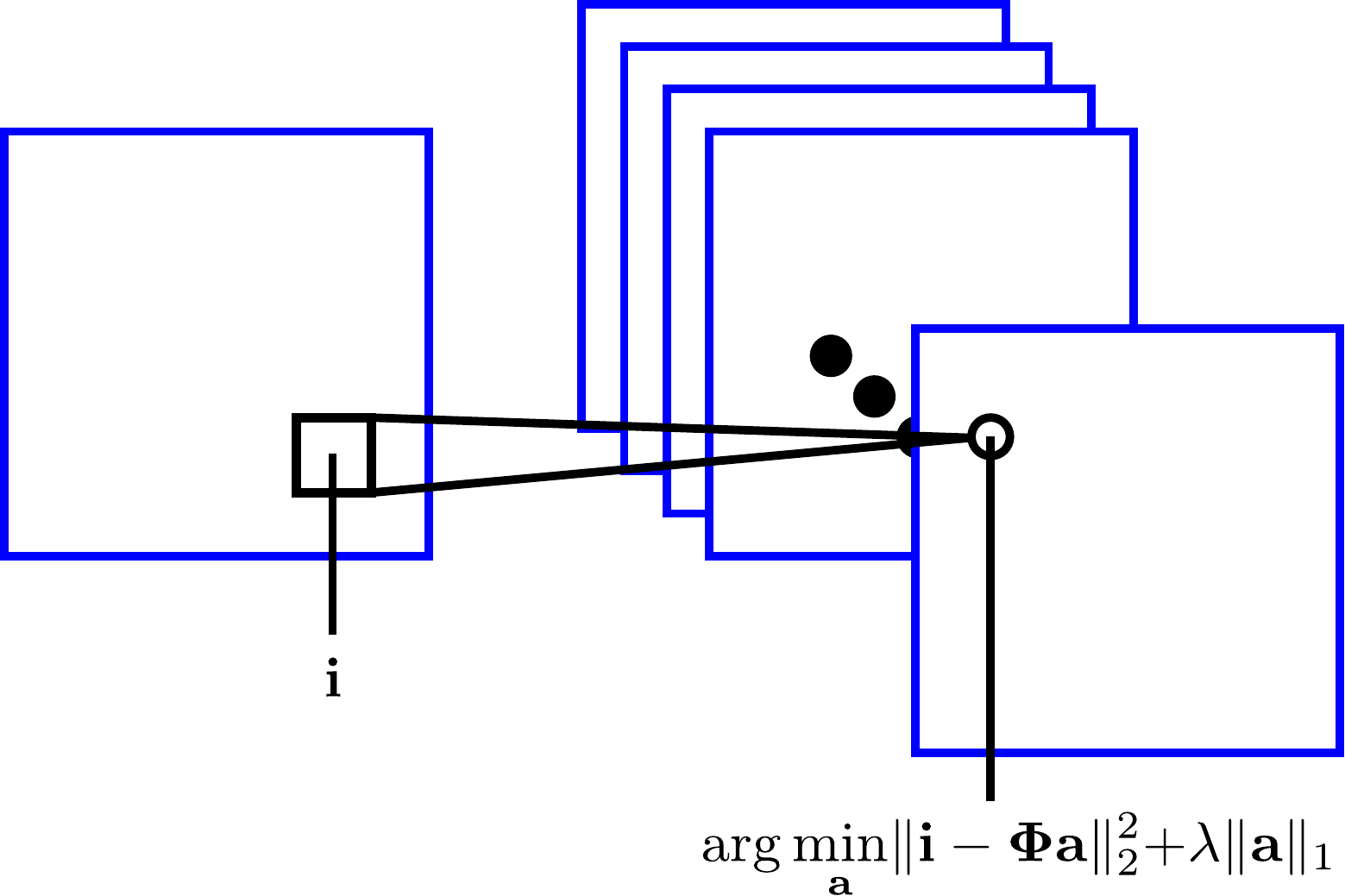}
		\caption{Sparse Coding with Sliding Window}
	\end{subfigure}
	\caption{\textbf{Adding Convolution Principles to Sparse Coding.} The vector $\mathbf{i}$ is the input gathered via the sliding window of convolution. The function $\boldsymbol{\sigma}$ is the ReLU activation function. The parameters of the two models were previously described. The sheets on the right of each image represents the number of filters $w_i$ or basis functions $\mathbf{\Phi}_i$. (a) Convolution with ReLU activation typically performed in a convolutional layer of a convolutional neural network. (b) Sparse Coding performed on inputs gathered from a sliding window. If the basis function matrix $\mathbf{\Phi}$ has orthogonal columns and $\mathbf{b}$ is a vector of all $-\lambda$, then the transformation in (a) is equivalent to that in (b).}
	\label{fig:convsc}
\end{figure}

Indeed, several have extended the usual sparse coding model into a convolutional one \citep{bristow:cvpr13,wohlberg:icassp14}, but most change the image model from a linear combination of basis functions to a sum of filters convolved with sparse feature maps. While this approach seems the natural extension of adding convolution to sparse coding, the image model loses its original probabilistic interpretation. Instead, the spatial consolidation component of convolution can be added to sparse coding by grouping together input vectors near to one another in space. Sparse coding can then be performed on the consolidated input vectors. This amounts to making a new set of inputs $\mathbf{a}_{\textrm{L1cons}}$ from the original sparse coefficients $\mathbf{a}_{\textrm{L1}}$. The sparse coefficients $\mathbf{a}_{\textrm{L1}}$ across space can be thought of as a feature map with dimensions $NxNxB$ where $N$ is the square-root of the number of input vectors and $B$ is the number of basis functions. The following layer then slides a $MxMxB$ window across the $NxNxB$ input feature map to gather its input vectors $\mathbf{a}_{\textrm{L1cons}}$ which may be written as
\begin{align}
\mathbf{a}_{\textrm{L1cons},i,j} &= 
\begin{bmatrix}
\mathbf{a}_{\textrm{L1},i,j} \\
\mathbf{a}_{\textrm{L1},i+1,j} \\
\vdots \\
\mathbf{a}_{\textrm{L1},(i+M-1),(j+M-1)}
\end{bmatrix},
\end{align}

\noindent for $0 \leq i \leq 2N - M$ and $0 \leq j \leq 2N - M$ where $\mathbf{a}_{\textrm{L1},i,j}$ is a set of $B$ sparse coding responses where each is arranged in a feature map indexed by $i$ and $j$. The stride shown is one. The difference between these two approaches being that instead of filtering the image with a linear filter and applying a ReLU nonlinearity, this approach performs the forward transform of sparse coding on the input (see figure \ref{fig:convsc}).

This approach has the advantage that it has an explicit underlying probabilistic interpretation. In other words, the model explicitly attempts to understand the distribution of its inputs. Convolutional neural networks also have a probabilistic interpretation, as shown by the derivation of its forward transform in section \ref{sec:cnnftder}. \emph{In fact, if the sparse coding basis functions are assumed to be orthogonal the model described here is exactly a convolutional neural network without pooling or normalization}. However, the forward transform of sparse coding does not require its basis functions to be orthogonal, so the model described here may do a better job at representing its input given its prior expectation on images. This along with maintaining a sparse prior may be essential to combating the issue of fooling in convolutional neural networks \citep{szegedy:arxiv13,nguyen:cvpr15}. One difficulty in this approach is that the forward transform of sparse coding contains nonlinear couplings with no closed-form solution, so approximating the solution to the convex optimization problem is more expensive then a simple ReLU of filtered input plus bias. Next we will see why maintaining a sparse prior and performing the full sparse inference (LASSO) transformation may be worth the trouble.

\section{Alternative Forward Transformations}
Viewing forward transforms as solutions to optimization problems introduces a rationale for the transformation and suggests possible alternatives. Rather than modeling neural computation as a nonlinear function of linearly filtered input, a model may derive its responses by solving a problem that the brain may solve. For the class of regularized least-squares loss functions, the rationale may be thought of as the need for forming a neural representation of the input (i.e. $i = \mathbf{\Phi} \mathbf{a}$) given a constraint on how the basis functions should be integrated (the regularization penalty). While this may not be the neural representation developed by the brain, it serves as a principled starting point. In addition to logically addressing the need for representation, another motivating point is to take inspiration from the brain in that sparse neural firing is observed in some contexts in the visual cortex \citep{willmore:jnp11}. This makes the L1 regularization term in sparse coding desirable, but other penalties may also be helpful. Here, we first consider the effect of constraining the bias weights of a convolutional neural network to maintain a sparse prior under the sparse coding interpretation proposed here. Next, we look at the effect of a L2 regularization term on the least-squares coefficients to gain insight as to why the orthogonality assumption under the sparse coding interpretation may hinder convolutional neural networks. Finally, we consider L0 regularization.

\subsection{Maintaining a Sparse Prior}\label{sec:mainprior}
In section \ref{sec:cnnftder} we saw that the convolutional neural network forward transformation can be derived by non-negative orthogonal sparse coding with an exponential prior parameter $\mu_i$ for each sparse coding coefficient. The choice of the parameters $\lambda_i$ determine both the bias weights and exponential prior parameters $\mu_i$. The importance of the sparse exponential prior is now evident from a function approximation interpretation of neural networks \citep{cybenko:mcss89,hanin:math19}. If the $\lambda_i$ are chosen to maximize classification accuracy, as in a convolutional neural network, the $\lambda_i$ can take on any value necessary to approximate a function defined by a set of training examples. However, if in addition to maximizing classification accuracy the $\lambda_i$ are chosen to maintain the prior distribution (positive $\mu_i$) as well as be sparse (relatively large $\lambda_i$), the function approximation will be biased towards the sparse prior. While the input data will not be fit perfectly, a perfect fit of the finite input data set is probably an example of overfitting which may be the case in convolutional neural networks. A sparse prior over the output neurons may help the model generalize, avoid overfitting, and properly integrate image features. Therefore, training a convolutional network with a constraint such that the bias neurons must be positive and relatively large (exact constraint must be determined experimently) is one alternative forward transformation under the sparse coding interpretation proposed in this work. The bias neuron weights might also be treated as constant hyperparameters (perhaps all with the same value) chosen a priori as in traditional sparse coding.

The next question evident from this claim is: why would a sparse prior promote proper feature integration (and thus generalization)? The answer to this question is explained by the inductive inference mechanism of sparse coding described by \cite{bowren:arxiv21}. As the regularization coefficient of sparse coding increases, fewer basis functions must reconstruct the input image. These reconstructions have higher error compared to sparse coding with a smaller regularization coefficient, but this error allows the basis functions to introduce their own information into the image representation. \cite{bowren:arxiv21} showed that in a hierarchical vision model when a small spatial region of the model's V1 complex cell layer responses was deleted, the missing image information could be inferred with 8x overcomplete non-negative sparse coding (see figures 14 and 15 of their work). In fact, as the value of the regularization coefficient increased, the amount of information able to be inferred also increased. The implication is that, given a sufficiently overcomplete representation, large regularization coefficients (and their corresponding priors with high kurtosis) help explain away artefactual information and give the model an understanding of how the basis functions should be integrated to represent image information. Given that the mechanism described by \cite{bowren:arxiv21} was studied in non-orthogonal non-negative sparse coding, the general (non-orthogonal) sparse coding transformation may be needed.

\subsection{Ridge Regression and its Connection to LASSO}
If we change the sparse coding exponential prior to a Gaussian prior with zero mean and variance $\frac{1}{\mu}$ (and the noise variance is still $\sigma^2$) we obtain the loss function of ridge regression instead of the L1-regularized least-squares problem:
\begin{align}
f_{rr}(\mathbf{a}) &= \frac{1}{2} \left\Vert \mathbf{i} - \mathbf{\Phi} \mathbf{a} \right\Vert_2^2 + \frac{\lambda}{2} \left\Vert \mathbf{a} \right\Vert_2^2
\end{align}

\noindent where $\lambda = \frac{\sigma^2}{\mu}$. The L2-penalty term might be thought of as an approximation of the L1-penalty term (just as the L1-penalty is an approximation of the L0-penalty). This approximation is not ideal, nor necessary, but the problem has a simple closed-form solution and helps provide insight into the issue of the orthogonality assumption. If we take the gradient of $f_{rr}$ and simplify we get
\begin{align}
\nabla_\mathbf{a} f_{rr}(\mathbf{a}) &= \nabla_\mathbf{a} \left( \frac{1}{2} \left\Vert \mathbf{i} - \mathbf{\Phi} \mathbf{a} \right\Vert_2^2 + \frac{\lambda}{2} \left\Vert \mathbf{a} \right\Vert_2^2 \right) \\
&= \mathbf{\Phi}^{\textrm{T}} \left( \mathbf{\Phi} \mathbf{a} - \mathbf{i} \right) + \lambda \mathbf{a} \\
&= \mathbf{\Phi}^{\textrm{T}} \mathbf{\Phi} \mathbf{a} - \mathbf{\Phi}^{\textrm{T}} \mathbf{i} + \lambda \mathbf{a} \\
&= (\mathbf{\Phi}^{\textrm{T}} \mathbf{\Phi} + \lambda \mathbf{I}) \mathbf{a} - \mathbf{\Phi}^{\textrm{T}} \mathbf{i} \\
\implies \mathbf{H}_{rr}(\mathbf{a}) &= \mathbf{\Phi}^{\textrm{T}} \mathbf{\Phi} + \lambda \mathbf{I}.
\end{align}

\noindent The Hessian $\mathbf{H}_{rr}$ is positive semi-definite, so the problem has a unique solution. The closed-form solution is obtained by setting the gradient to zero:
\begin{align}
0 &= (\mathbf{\Phi}^{\textrm{T}} \mathbf{\Phi} + \lambda \mathbf{I}) \mathbf{\hat{a}} - \mathbf{\Phi}^{\textrm{T}} \mathbf{i} \\
&\implies \mathbf{\Phi}^{\textrm{T}} \mathbf{i} = (\mathbf{\Phi}^{\textrm{T}} \mathbf{\Phi} + \lambda \mathbf{I}) \mathbf{\hat{a}} \\
&\implies \mathbf{\hat{a}} = (\mathbf{\Phi}^{\textrm{T}} \mathbf{\Phi} + \lambda \mathbf{I})^{-1} \mathbf{\Phi}^{\textrm{T}} \mathbf{i}\label{eq:rrsol}.
\end{align}

\noindent Notice that the solution is only a linear function of the basis function matrix $\mathbf{\Phi}$ and regularization coefficient $\lambda$. Also notice that if the basis functions have orthogonal columns, the solution is 
\begin{align}
\mathbf{\hat{a}} &= (\mathbf{I} + \lambda \mathbf{I})^{-1} \mathbf{\Phi}^{\textrm{T}} \mathbf{i} \\
&= \frac{1}{\lambda + 1} \mathbf{\Phi}^{\textrm{T}} \mathbf{i}.\label{eq:rrorthsol}
\end{align}

\noindent \emph{Interestingly, the solution for orthogonal ridge regression is the same as that for orthogonal LASSO except for the soft-thresholding operation which is replaced by the scaling term $\frac{1}{\lambda + 1}$}. The scaling term may be viewed as the analog in ridge regression of the soft-thresholding operation in LASSO. The important finding here is that the orthogonality assumption makes LASSO and ridge regression roughly equivalent for small $\lambda$, therefore the nonlinearity from the forward transform of sparse coding is minimal under the orthogonality assumption. We can get a sense of the precision lost when making the orthogonality assumption by looking at the solution to ridge regression in equation \ref{eq:rrsol}. The matrix $(\mathbf{\Phi}^{\textrm{T}} \mathbf{\Phi} + \lambda \mathbf{I})^{-1}$ is approximated by $\frac{1}{\lambda + 1}$ which is unlikely to be a good approximation. There is an analog in LASSO that shows its loss of precision under the orthogonality assumption, but the analog is nonlinear and not easily written since there is no general closed-form solution.

\subsection{L0-Regularized Least Squares}\label{sec:l0rls}
Regularization with the L0 norm is generally considered the best principled method of enforcing sparsity in least-squares, but is not often incorporated into sparse coding because it cannot be easily computed (the number of non-zero elements must be counted). \cite{rehn:jcn07} provided a method of approximating the L0-norm, and performed L0-regularized sparse coding which obtained results similar to highly-overcomplete sparse coding \citep{olshausen:wavelets09,olshausen:hvei13}. However, under the orthogonality assumption the L0-norm can be computed exactly with a closed-form solution. As described by \cite{schutze:ieeeci16}, the solution is to keep the $\lfloor \lambda \rfloor$ largest values of $| \mathbf{\Phi}^{\textrm{T}} \mathbf{i} |$ with the rest being set to zero. While the orthogonality assumption is problematic for the reasons previously discussed, approximations of L1- and L0-regularized least-squares are computationally expensive. For L1-regularized least squares, the ReLU of $\mathbf{\Phi} \mathbf{a} - \boldsymbol{\lambda}$ may be motivated by the need for an efficient approximation. With this rationale, orthogonal L0-regularized least-squares may replace orthogonal L1-regularized least-squares. The activation function of this method would be similar to a ReLU, however only setting values to zero if the number of non-zero coefficients $\lfloor \lambda \rfloor$ is exceeded. The activation function may be written as 
\begin{align}
	h_{x_{\lambda thlargest}}\left(x\right) &=
	\begin{cases} 
		0 & x < x_{\lambda thlargest} \\
		x & otherwise,
	\end{cases}
\end{align}

\noindent where $x_{\lambda thlargest}$ is the $\lambda th$ largest element of $\mathbf{x}$. It is important to note that $\lambda$ cannot be varied to a vector $\boldsymbol{\lambda}$ because the L0 norm is only measuring sparseness in $\mathbf{a}$ rather than the distribution of $a_i \forall i$. The L0 norm would have to measure sparseness across the firing of each individual $a_i$ over the set of input images.

\section{Discussion}

This paper provided an efficient solution to L1-regularized orthogonal sparse coding, displayed the orthogonal sparse coding basis functions, provided a sparse coding derivation for the forward transform of neural networks incorporating ReLU activation, extended the derivation to a complete convolutional neural network without pooling or normalization, and suggested improvements that may make for more robust neural networks that are resistant to fooling like that of \cite{szegedy:arxiv13} and \cite{nguyen:cvpr15}. To our knowledge, aside from the links between sparse coding and convolutional neural networks provided by \cite{fawzi:ijcv15} and \cite{papyan:jmlr17}, no comprehensive derivation of a convolutional neural network from the first principles of sparse coding has been published. Here is provided a derivation of both the forward transform (ReLU of filtered input) and loss function of a convolutional neural network from a slight modification to the sparse coding model of \cite{olshausen:vr97}. Future work may investigate an implementation of the full model and compare its performance with convolutional neural networks on image classification and its ability to be fooled by computer-generated images.

The orthogonality assumption of sparse coding basis functions is unlikely to hold in general, but it was found that optimizing the model without orthogonalizing the basis functions after each update was not strictly necessary; the model could still be trained with a smaller learning rate, albeit at a slower rate. Orthogonalization kept the sparse coding basis functions stable (real values) during training with a larger learning rate, and the constraint resulted shorter training times. This was interesting because the result implies that other loss functions, like the cross entropy loss function, might integrate with orthogonal sparse coding without enforcing orthogonality in the basis functions. When the hyperparameter $\lambda$ is varied for each sparse coefficient $a_i$ (giving parameters $\lambda_i$), the non-negative (more biologically plausible) version of the forward transform is equivalent to the forward transform of a convolutional neural network (see section \ref{sec:cnnftder}), so the ability to train the model without orthogonalizing the basis functions may be thought of as the reason a convolutional neural network does not require orthogonalization under this sparse coding interpretation. Interestingly, the orthogonality assumption allowed for the sparse inference problem to be solved exactly with a closed-form solution, leaving only one optimization problem necessary: deriving the basis functions to estimate the distribution of natural images. This convenient engineering abstraction speeds up the learning process and facilitates directly learning model parameters (the basis functions) in hierarchical sparse coding models. For example, gradient computation can be performed on the closed-form solution to the sparse inference problem whereas hierarchical sparse coding without the orthogonality assumption cannot directly compute the gradient of the optimization problem in equation \ref{eq:sparseinf}. Instead, the sparse coefficients must be inferred before taking a gradient step to minimize the overall loss function, but the nonlinearity of the closed-form solution is not incorporated into the gradient.

The orthogonal sparse coding model derived here differed from that of \cite{schutze:ieeeci16} in that it optimized the L1-regularized least-squares loss function rather the L0-regularized form. While the L0 norm is generally considered a better measure of sparsity, the solution to the L0-regularized problem requires searching $\mathbf{\Phi}^{\textrm{T}} \mathbf{i}$ for the $\lfloor\lambda\rfloor$ values with the largest magnitudes and setting the rest to zero. The L1-regularized problem by contrast has a closed-form solution that does not require finding the coefficients with the largest magnitudes, and can be written as $\mathbf{ReLU}(\mathbf{\Phi}^{\textrm{T}} \mathbf{i} - \lambda)$. The difference of regularization penalties likely resulted in the difference between the basis functions shown in figure \ref{fig:orthscvsscvsicaorth} and those of \cite{schutze:ieeeci16}. The Gaussian envelope of the Gabor functions learned with the L0 regularization penalty mostly encompassed the entire receptive field or only a small region whereas L1-regularized orthogonal sparse coding Gabor functions usually had a Gaussian envelope of an intermediate size (similar to regular sparse coding and ICA). Some of the orthogonal sparse coding basis functions learned here resembled the double-Gabor functions described by \cite{saremi:nc13}. For example, the basis functions in row 3 column 4 and row 8 column 7 of figure \ref{fig:orthscvsscvsicaorth} resembled double-Gabor functions with a 90-degree phase-shift midway. More such double-Gabor functions are apparent in figure \ref{fig:orthscbases}.

Also interesting was the trend of L1-regularized orthogonal sparse coding to find fewer high spatial frequency basis functions with fewer total basis functions (see figure \ref{fig:orthscbases}). Orthogonal sparse coding was performed here on 12x12 image patches, so 144 basis functions corresponds to a complete representation. However, a complete representation, and some slightly under complete representations (e.g. 121) produced several high spatial frequency basis functions (see row 2 column 1 of figure \ref{fig:orthscbases144}) that likely did not significantly contribute to the image reconstruction. The discovery of these basis functions can be attributed to the learning of an orthogonal basis. There can be at most $n$ linearly-independent vectors in a $n$-dimensional vector space, so there can be at most $144$ uncorrelated vectors (the orthogonal sparse coding constraint enforced only first-order statistics to be zero). While high spatial frequency basis functions were still found within a vector space of $121$ and $144$ (see figures \ref{fig:orthscbases121} and \ref{fig:orthscbases144}), these can be attributed to the choice of the adjustable regularization coefficient $\lambda$ (0.1). This objective can also be thought of as ICA with Gaussian noise and the addition of a hyperparameter (regularization coefficient) choosing the kurtosis of the Laplacian prior over the independent components \citep{hyvarinen:ncomputing98,hyvarinen:nc99}. Usually, ICA without noise is formulated without the ability to change the kurtosis of the Laplacian prior, so this approach may be thought more of as orthogonal sparse coding. The difference in basis functions learned by ICA and orthogonal sparse coding in figures \ref{fig:orthscvsscvsicaorth} and \ref{fig:orthscvsscvsicaica} may be attributed the choice of regularization coefficient for orthogonal sparse coding. Another difference between orthogonal sparse coding and ICA is that orthogonal sparse coding does not require a square basis function matrix which allows the model to compress its input, though expanding should be superfluous. In this interpretation, ICA with noise is a special case of orthogonal sparse coding and has the same closed-form solution for a particular Laplacian hyperparameter.

Since orthogonal sparse coding with L1 regularization had an exact closed-form solution which was a function of $\mathbf{\Phi}^{\textrm{T}} \mathbf{i}$, a direct comparison with neural networks could be made. In fact, the forward transformation of orthogonal sparse coding turned out to be a neural network with a special activation function: the soft-threshold function, given by equation \ref{eq:soft-threshold2}. While most neural networks motivate their choice of activation function by forming nonlinear decision boundaries (the logistic function and hyperbolic tangent function) or speeding up learning and avoiding pitfalls like vanishing gradients (ReLU), the activation function of this model interpreted as a neural network is motivated by the orthogonal version of the sparse inference problem (see equation \ref{eq:sparseinf}). A visual examination of the model's activation function plotted in figure \ref{fig:soft-threshold} bears a resemblance to the ReLU activation function (see the graph on the right of figure \ref{fig:nnscvscnn}) with the positive portion shifted to the right by the regularization coefficient $\lambda$ and reflected over the x-axis and y-axis. Conveniently, a non-negative formulation of the usual sparse coding model exists \citep{hoyer:nnsp02}, and adding the constraint to orthogonal sparse coding was trivial (see section \ref{sec:nnsc}). The result of this addition was the removal (setting to zero) of the portion of the solution to the left of $\lambda$. After examining the plot of the non-negative solution (see the graph on the left side of figure \ref{fig:nnscvscnn}), the connection to the ReLU function was clear: \emph{the activation function of non-negative orthogonal sparse coding is the ReLU function shifted to the right by the regularization coefficient $\lambda$, and the overall forward transformation is the same as a convolutional neural network with a constant bias neuron value of $-\lambda$ for all the neurons}.

With the connection of non-negative orthogonal sparse coding to the ReLU function and convolutional neural networks clear, the derivation of the convolutional neural network forward transform required little extra work. The hyperparameter $\lambda$ for each exponential distribution of each sparse coefficient $a_i$ was allowed to vary as the parameters $\lambda_i$ in order to let each coefficient $a_i$ (output neuron) have its own bias neuron input (with value $-\lambda_i$) instead of having one bias neuron for all the coefficients. The model of \cite{karklin:nc05} also allowed different $\lambda_i$ for each coefficient $a_i$ (computed as a nonlinear function of latent variables), but here the change was made to make the connection with convolutional neural networks. The derivation proceeded the same way as that for non-negative orthogonal sparse coding (see section \ref{sec:cnnftder}). The difference between the findings of \cite{fawzi:ijcv15} and \cite{papyan:jmlr17} and this derivation of the convolutional neural network forward transform was that \cite{fawzi:ijcv15} and \cite{papyan:jmlr17} provided links to closely related problems. Both provided models with sparse dictionaries that directly applied the basis functions as filter vectors to the image \citep[e.g. solving the Sparse-Land model][with soft-thresholding of filter responses]{elad:tip06}. Here, the base model for this work was that of \cite{olshausen:nat96}. The prior (exponential) distribution parameter $\mu$ was varied for each coefficient $a_i$ in order for the solution to match the forward transform of a convolutional neural network exactly. This way, the output of a hierarchical orthogonal sparse coding model and a convolutional neural network can be compared more directly.

It is important to note that while the forward transform of non-negative orthogonal sparse coding with exponential parameters $\mu_i$ exactly matches that of a convolutional neural network, the probabilistic interpretation is lost when any hyperparameter $\lambda_i = \frac{\sigma^2}{\mu_i}$ is chosen to be negative because the Gaussian noise variance $\sigma^2$ and exponential prior parameters $\mu_i$ are defined as positive values for the prior distributions. Negative regularization coefficients $\lambda_i$ also remove the sparseness constraint of the loss function for the corresponding coefficients $a_i$ because the contribution to the loss function $\lambda_i a_i$ rewards large coefficients. So, a trained convolutional neural network with some negative and positive bias weights (positive and negative $\lambda_i$) chooses some neurons to be sparse and rewards others for large coefficients. Considering again the non-negative sparse coding model without varied $\mu_i$, the connection to convolutional neural networks still holds without bias weights and with a shifted ReLU, or a single bias weight ($-\lambda$) for all output neurons and a regular ReLU. However, the regularization coefficient $\lambda$ must be positive to maintain the probabilistic interpretation. In other words, the ReLU must only shift to the right.

The lack of a probabilistic interpretation in a convolutional neural network under this sparse coding interpretation may be problematic because the prior expectation in the model via the prior distribution (exponential distribution) is lost. This issue may contribute to fooling convolutional neural networks \citep{szegedy:arxiv13,nguyen:cvpr15} because the expectation of image structure may be poor (some negative $\lambda_i$) or completely absent (all negative $\lambda_i$). Given that the parameters $\lambda_i$ were chosen to maximize classification accuracy, and not preserve the probabilistic interpretation, this may be related to overfitting. Maximizing classification accuracy given a sparse prior may be thought of as a way of traversing the landscape of the loss function along representations that go together according to some logic (signals can usually be represented by a few underlying causes). Though one may argue that maximizing classification accuracy without a sparse prior inherently preserves a prior expectation, the prior may be a different one not characterized by this sparse coding interpretation. However, an algorithm like gradient descent need not find a representation with a prior expectation if there exists a solution with low error highly specific to the training set, but without the need for a set of rules for how representations should model images. One such model may be thought of as a model with many high-level features that are useful for signaling a certain label, but a disregard for how the features should be integrated together. Issues arising from an incorrect prior may work their way into convolutional neural networks given that the hierarchical non-negative orthogonal sparse coding model discussed here has the same loss function, discussed next.

When the coefficients $\mathbf{a}_{\textrm{L1}}$ and $\mathbf{a}_{\textrm{L2}}$ (see section \ref{sec:stacksparse}) are learned independently with the corresponding sets of regularization coefficients $\lambda_{\textrm{L1},i}$ and $\lambda_{\textrm{L2},i}$ via non-negative orthogonal sparse coding, but the basis function matrices $\mathbf{\Phi}_{\textrm{L1}}$ and $\mathbf{\Phi}_{\textrm{L2}}$ are learned via another loss function over all parameters, the cross-entropy loss function, the model is a convolutional neural network without convolution, pooling, or normalization (see section \ref{sec:addlogreg}). The cross-entropy loss function arises from multinomial logistic regression \citep{bishop:prml06}. The connection between logistic regression and sparse coding was made here by changing the minimum KL-divergence interpretation of \cite{olshausen:vr97} to minimize the KL-divergence of the conditional distribution of a target label given an image with the label $P(\mathbf{t} | \mathbf{i}_{\mathbf{t}})$ and its distribution under the model $P(\mathbf{t} | \mathbf{\theta}, \mathbf{i}_{\mathbf{t}})$ (where $\mathbf{\theta}$ is the set of model parameters) instead of the distribution of natural images $P(\mathbf{i})$ and its distribution under the model $P(\mathbf{i} | \mathbf{\theta})$. The overall model is therefore partly generative and partly discriminative (partly unsupervised and partly supervised); the sparse inference portion attempts to sparsely reconstruct images while the basis functions are optimized for distinguishing between images. While the model lacks convolution, it is trivial to add a convolution-like procedure where sparse inference replaces the dot product of a neural network (see section \ref{sec:sparseconv}). Previous methods of convolutional sparse coding modified the loss function to reconstruct images as a sum of filters convolved with sparse feature maps \citep{bristow:cvpr13,wohlberg:icassp14}, but this loses the original probabilistic interpretation of sparse coding and the associated inference capabilities described by \cite{bowren:arxiv21}. Instead, this work proposed gathering image patches via a sliding window (like in convolution), but performing sparse inference on each patch. \emph{This method is equivalent to a convolutional neural network without pooling or normalization}. Pooling may be connected to the model via undercomplete orthogonal sparse coding layers, but the rationale and connection to biology is not clear. Normalization on the other hand can be principled and biologically-motivated, but it is unclear on the surface whether the benefits from a normalization technique such as that of \citep{coen:plos12,coen:jov13} would improve the model's ability to find useful representations.

This understanding of convolutional neural networks in terms of sparse coding makes several potential improvements apparent. First, one change to attempt is to constrain the bias weights to be negative (positive $\lambda$; see previous discussion) in order to ensure that the probabilistic interpretation holds true. The magnitude of the bias weights should also be relatively large to benefit from inference in sparse coding as seen from reconstructions with large values of the regularization coefficient (see section \ref{sec:mainprior}), but the percise constraint must be determined experimentally by training on images. Given that convolutional neural networks give the same label for images when local features are scrambled \citep{brendel:arxiv19}, the prior representation may be suffering, and maintaining the sparse coding prior may help. Second, one can attempt to train a convolutional neural network with one (perhaps constant) negative bias weight (positive $\lambda$) for each layer. This is equivalent to hierarchical non-negative orthogonal sparse coding. A constant negative bias weight allows the researcher to choose the degree of sparsity in each layer (and favor the prior expectation over classifying power in an attempt to avoid overfitting), but a single bias weight optimized for image classification may suggest the best (perhaps biologically plausible) level of sparsity. However, one issue that remains in this approach is that there is no good reason to assume orthogonality in the model's basis functions. The third potential improvement is to run non-negative sparse coding without the orthogonality assumption. The solution to the sparse coding inference problem given by equation \ref{eq:sparseinf} is a nonlinear function of the input image $\mathbf{i}$ and basis function matrix $\mathbf{\Phi}$, but assuming orthogonal basis functions reduces the nonlinearity to the simplest case: a piecewise point-nonlinearity with two conditions (or one for non-negative sparse coding). Such a reduction in transformative power may be problematic for deriving the best model representation for computer vision. The loss of transformative power can be seen via the analog of ridge regression, discussed next.

Unlike LASSO, ridge regression has a closed-form solution without the orthogonality assumption. The solution is a linear function of $\mathbf{a}$ and $\mathbf{\Phi}$ given by equation \ref{eq:rrsol}. Under the orthogonality assumption, the solution is also a linear function, but a point-linear function (linear function of $\mathbf{\Phi}^{\textrm{T}} \mathbf{i}$) given by equation \ref{eq:rrorthsol}. It is likely the case that a similar reduction in transformative power occurs with LASSO which is what made it a point-nonlinear function. In fact, the orthogonality assumption almost reduces the orthogonal LASSO (orthogonal sparse inference) solution to the orthogonal ridge regression solution except for the rectification step in orthogonal LASSO replacing the $\frac{1}{\lambda + 1}$ linear scaling term in orthogonal ridge regression. In other words, it is likely that important transformations are lost under the orthogonality assumption. A similar effect is seen for orthogonal L0-regularized least-squares. As mentioned before, the solution to the L0-regularized orthogonal least-squares (sparse inference step of sparse coding, see section \ref{sec:l0rls}) is the vector $\mathbf{\Phi}^{\textrm{T}} \mathbf{i}$ with the $\lfloor \lambda \rfloor$ largest components maintained and the rest set to $0$. So once again, the solution is a point-nonlinear function of $\mathbf{\Phi}^{\textrm{T}} \mathbf{i}$. L0-regularized sparse coding may be explored to improve neural networks, but since solving the L0 norm is similar to solving the L1 norm for large $\lambda$ \citep{donoho:nas03,donoho:ieeeit01,elad:icip01,elad:ieeeit02}, the L1 approximation may be preferred for highly sparse solutions.

One last potential improvement for a convolutional neural network interpreted as a hierarchical sparse coding model is to learn all the sparse coefficients at once (previously mentioned for hierarchical sparse coding). A convolutional neural network can compute the values of its outputs at each layer with only the inputs to the layer (referred to as a feedforward neural network). It is possible that coefficients in the last layer should influence the value of the coefficients in all other layers, even the first layer, upon computing the entire set of outputs. Such a scheme is motivated by the feedback and feedforward connections in the brain. One such way to learn all coefficients at once is to optimize the sparse coding loss function for all sets of coefficients together instead of one set at a time. This way, dependencies between the coefficients of each layer may be learned. \cite{boutin:plos21} added this technique to sparse coding models with some success. The implication of even some of these improvements may be striking. Convolutional neural networks and/or hierarchical sparse coding models may be derived that have a richer understanding of image classes and how features should be integrated. More importantly, the theoretical framework here implies that convolutional neural networks under some constraints are hierarchical sparse coding models, and their image transformations are mathematically principled and to some extent biologically plausible given the correct modifications.

\section{Appendix A}\label{sec:appA}

\subsection{Sparse Coding Subdifferential Cases}

The derivative of the absolute value function is undefined at 0, but defined everywhere else. An equality symbol can be written when the derivative is defined instead of the $\in$ symbol because the subdifferential set contains one element, i.e. the derivative. For the case $a_i > 0$, the equation is
\begin{align}
	&\hat{a}^{orth}_i = \mathbf{\Phi}_i^{\textrm{T}} \mathbf{i} - \lambda \\
	&\implies \mathbf{\Phi}_i^{\textrm{T}} \mathbf{i} - \lambda > 0 \\
	&\implies \mathbf{\Phi}_i^{\textrm{T}} \mathbf{i} > \lambda.
\end{align}

\noindent For the case $a_i < 0$, the equation takes the form
\begin{align}
	&\hat{a}^{orth}_i = \mathbf{\Phi}_i^{\textrm{T}} \mathbf{i} + \lambda \\
	&\implies \mathbf{\Phi}_i^{\textrm{T}} \mathbf{i} + \lambda < 0 \\
	&\implies \mathbf{\Phi}_i^{\textrm{T}} \mathbf{i} < \lambda.
\end{align}

\noindent For the case $a_i = 0$, the subdifferential set $\partial \left| a_i \right|$ no longer contains one element, but can take on any value in the range $\left[-\lambda, \lambda\right]$. The equation then takes on the form
\begin{align}
	&0 \in \mathbf{\Phi}_i^{\textrm{T}} \mathbf{i} - \left[-\lambda,\lambda\right] \\
	&\implies \mathbf{\Phi}_i^{\textrm{T}} \mathbf{i} \in \left[-\lambda,\lambda\right]
\end{align}

\noindent Each case put a constraint on the term $\mathbf{\Phi}_i^{\textrm{T}} \mathbf{i}$ which allows the solution to be written as a piecewise function changing for different values of $\mathbf{\Phi}_i^{\textrm{T}} \mathbf{i}$:
\begin{align}
	\hat{a}^{orth}_i = s\left(\mathbf{\Phi}_i^{\textrm{T}} \mathbf{i}\right) =
	\begin{cases} 
		\mathbf{\Phi}_i^{\textrm{T}} \mathbf{i} - \lambda & \mathbf{\Phi}_i^{\textrm{T}} \mathbf{i} > \lambda \\
		0 & \left| {\mathbf{\Phi}_i^{\textrm{T}} \mathbf{i}} \right| \leq \lambda \\
		\mathbf{\Phi}_i^{\textrm{T}} \mathbf{i} + \lambda & \mathbf{\Phi}_i^{\textrm{T}} \mathbf{i} < -\lambda.
	\end{cases}
\end{align}

\section{Appendix B}\label{sec:appB}

\subsection{Non-negative Sparse Coding Subdifferential Cases}

The subdifferential set of the right most term of equation \ref{eq:nnsol} has only one subderivative for $a_i > 0$, but does not exist when $a_i < 0$. For the case $a_i > 0$ an equality symbol can be written:
\begin{align}
	&\hat{a}^{orth}_i = \mathbf{\Phi}_i^{\textrm{T}} \mathbf{i} - \lambda \\
	&\implies \mathbf{\Phi}_i^{\textrm{T}} \mathbf{i} - \lambda > 0 \\
	&\implies \mathbf{\Phi}_i^{\textrm{T}} \mathbf{i} > \lambda.
\end{align}

\noindent For the case $a_i = 0$, the subdifferential set takes on any value in the range $\left[-\infty,\lambda\right]$, with the bound $-\infty$ appearing because the log function is undefined at 0.
\begin{align}
	&0 \in \mathbf{\Phi}_i^{\textrm{T}} \mathbf{i} - \left[-\infty,\lambda\right] \\
	&\implies \mathbf{\Phi}_i^{\textrm{T}} \mathbf{i} \in \left[-\infty,\lambda\right].
\end{align}

\noindent While this yields the solution for $\mathbf{\Phi}_i^{\textrm{T}} \mathbf{a} < 0$ as well as $0 \leq \mathbf{\Phi}_i^{\textrm{T}} \mathbf{a} < \lambda$, formally the subdifferential set cannot be computed where the negative log-likelihood function is undefined. A small probability of $b$ can replace 0 for negative coefficients, so that the log function exists. In effect, the subdifferential set for the negated last term of equation \ref{eq:nnsol} and $\mathbf{\Phi}_i^{\textrm{T}} \mathbf{a}$ take on values in the range $\left[0,\lambda\right]$ for $a_i = 0$. For $\mathbf{\Phi}_i^{\textrm{T}} \mathbf{a} < 0$, recall from section \ref{sec:nnsc} that for any value of $a_i < 0$, any other value is at least as good or better (as long as $P(a_i) \geq b$ for the other value) and that the value $a_i = 0$ is strictly better than any $a_i < 0$ (as long as $b < \exp\left(-\frac{1}{2} \Vert \mathbf{i} \Vert_2^2\right)$). Finally, since both terms of the loss function in equation \ref{eq:nnsol} decrease as $a_i$ decreases for $\mathbf{\Phi}_i^{\textrm{T}} \mathbf{i} < 0$, the minimum is $a_i = 0$. The solution takes the form
\begin{align}
	\hat{a}^{orth}_i = s_{nn,\lambda}(\mathbf{\Phi}_i^{\textrm{T}} \mathbf{i}) &=
	\begin{cases} 
		\mathbf{\Phi}_i^{\textrm{T}} \mathbf{i} - \lambda & \mathbf{\Phi}_i^{\textrm{T}} \mathbf{i} > 0 \\
		0 & \textrm{otherwise}.
	\end{cases}
\end{align}

\section{Acknowledgments}

The author would like to thank his Lord Jesus Christ for the wisdom obtained for this work after much prayer. The author would like to thank Odelia Schwartz for her helpful comments about the paper. The author would like to thank Bruno Olshausen for his discussion about hierarchical sparse coding and Manohar Murthi for his discussion about solving the sparse inference problem with one basis function via the subdifferential set. This material is based upon work supported by the National Science Foundation Graduate Research Fellowship Program under Grant No. 1451511. Any opinions, findings, and conclusions or recommendations expressed in this material are those of the author(s) and do not necessarily reflect the views of the National Science Foundation.

\end{document}